\documentclass[10pt,twocolumn,letterpaper]{article}

\usepackage{wacv}
\usepackage{times}
\usepackage{epsfig}
\usepackage{graphicx}
\usepackage{mathrsfs,amsmath}
\usepackage{amssymb}
\usepackage{mathtools}
\usepackage{dsfont}
\usepackage{multirow}
\usepackage{hhline}
\usepackage{algorithm,algpseudocode,amsthm,mathrsfs,subcaption}

\wacvfinalcopy % *** Uncomment this line for the final submission

\def\wacvPaperID{867} % *** Enter the wacv Paper ID here

\def\r{\boldsymbol{r}}
\def\x{\boldsymbol{x}}
\def\w{\boldsymbol{w}}
\def\m{\boldsymbol{m}}
\def\v{\boldsymbol{v}}
\def\g{\boldsymbol{g}}
\def\h{\boldsymbol{h}}

\def\ie{\textit{i.e.}}
\DeclareMathOperator*{\argmax}{arg\,max}
\DeclareMathOperator*{\argmin}{arg\,min}

\newcommand{\overbar}[1]{\mkern 1.5mu\overline{\mkern-1.5mu#1\mkern-1.5mu}\mkern 1.5mu}

\ifwacvfinal\fi
\begin{document}

%%%%%%%%% TITLE
\title{SmoothFool: An Efficient Framework for Computing Smooth Adversarial Perturbations}

% Authors at the same institution
%\author{First Author \hspace{2cm} Second Author \\
% Authors at different institutions
\author{Ali Dabouei, Sobhan Soleymani, Fariborz Taherkhani, Jeremy Dawson, Nasser M. Nasrabadi\\
West Virginia University\\
~\\
{\tt\small \{ad0046, ssoleyma\}@mix.wvu.edu, fariborztaherkhani@gmail.com,}\\
{\tt\small\{nasser.nasrabadi, jeremy.dawson\}@mail.wvu.edu }}

\maketitle
\ifwacvfinal\thispagestyle{empty}\fi

%%%%%%%%% ABSTRACT
\begin{abstract}
Deep neural networks are susceptible to adversarial manipulations in the input domain. %carefully manipulated input samples called adversarial examples. 
The extent of vulnerability has been explored intensively in cases of $\ell_p$-bounded and $\ell_p$-minimal adversarial perturbations. However, the vulnerability of DNNs to adversarial perturbations with specific statistical properties or frequency-domain characteristics has not been sufficiently explored. In this paper, we study the smoothness of perturbations and propose SmoothFool, a general and computationally efficient framework for computing smooth adversarial perturbations. Through extensive experiments, we validate the efficacy of the proposed method for both the white-box and black-box attack scenarios. In particular, we demonstrate that: (i) there exist extremely smooth adversarial perturbations for well-established and widely used network architectures, (ii) smoothness significantly enhances the robustness of perturbations against state-of-the-art defense mechanisms, (iii) smoothness improves the transferability of adversarial perturbations across both data points and network architectures, and (iv) class categories exhibit a variable range of susceptibility to smooth perturbations.  Our results suggest that smooth APs can play a significant role in exploring the vulnerability extent of DNNs to adversarial examples. % which lie beyond the conventional $\ell_p$-norm constraints.% 
The code is available at \textit{\textbf{ https://github.com/alldbi/SmoothFool}}
\end{abstract}

%%%%%%%%% BODY TEXT
\section{Introduction}

Despite revolutionary achievements of deep neural networks (DNNs) in many computer vision tasks \cite{krizhevsky2012imagenet, zhang2016augmenting}, carefully manipulated input samples, known as \textit{adversarial examples}, can fool learning models to confidently make wrong predictions \cite{szegedy2013intriguing}. Adversarial examples are potential threats to almost all applications of machine learning \cite{carlini2018audio, grosse2017adversarial, huang2017adversarial}, but the case is more severe in the context of computer vision, particularly, due to the complexity of tasks \cite{shafahi2018adversarial}, huge cardinality of input spaces \cite{tramer2018ensemble}, and sensitivity of applications \cite{fischer2017adversarial, hendrik2017universal, Xie2017adversarial, song2018physical}. Analyzing DNNs as differentiable transfer functions have led to substantial studies exploring embedding spaces and their characteristics in regard to training paradigms. However, the adversarial behavior has highlighted the  importance of studying the topology of decision boundaries and their properties in high dimensional data spaces \cite{shafahi2018adversarial, hein2017formal}.
\begin{figure}
    \centering
    \includegraphics[width=210pt]{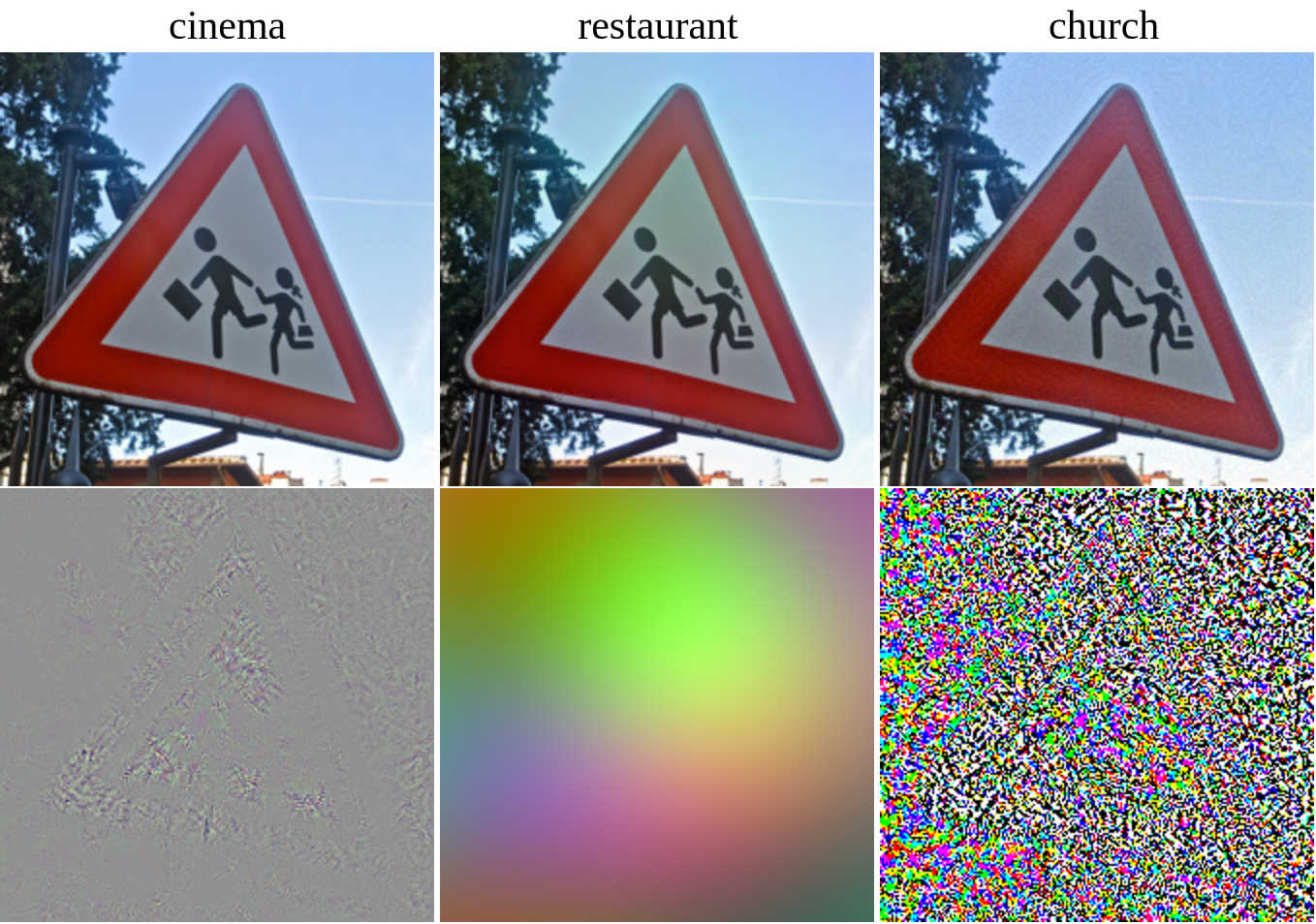}
    \caption{Comparing smooth APs with conventional APs. Each column from left to right shows the adversarial image and the corresponding APs computed by DeepFool \cite{moosavi2016deepfool}, SmoothFool ($\sigma_{\g}=75$) and IGSM \cite{kurakin2016adversarial}, respectively, on ResNet-101 \cite{he2016deep}. The predicted label for each image is depicted above the column. Perturbations are magnified for a better visibility.
    }
    \label{fig:firstexample}
    \vspace{-15pt}
\end{figure}
Considering a \textit{white-box} scenario where the network architecture and all its parameters are known, several approaches (attacks) have been proposed to explore the robustness of decision boundaries in the presence of $\ell_p$-bounded \cite{szegedy2013intriguing, goodfellow2015explaining, kurakin2016adversarial} and $\ell_p$-minimal \cite{moosavi2016deepfool, carlini2017towards, papernot2016limitations, su2017one, modas2018sparsefool} adversarial perturbations (APs). However, the vulnerability of DNNs to APs with specific statistical properties or frequency-domain characteristics, which lie beyond the conventional $\ell_p$-norm constraints, has remained less explored.

In this study, we seek to explore the landscape of robustness of DNNs to APs with modified frequency-domain characteristics. Specifically, we focus on smooth APs due to several advantages they offer compared to the conventional APs. First, they are more physically realizable than non-smooth APs since printing devices are critically less accurate in capturing high frequency structures due to the sampling noise \cite{sharif2016accessorize}. Also, severe differences between adjacent pixels in the printed adversarial examples are unlikely to be accurately captured by cameras due to their low-pass spatial frequency response \cite{jan2019connecting}. Second, the high-frequency structure of conventional APs has provoked an intensive adoption of explicit \cite{moosavi2018divide, Liao2018defense, prakash2018deflecting} and implicit \cite{tramer2018ensemble, madry2018toward, samangouei2018defense} denoising methods to mitigate the adversarial effect. However, we demonstrate that a slight modification of local statistics of APs causes a vital failure of state-of-the-art defenses. Third, smoothness significantly enhances the transferability of APs across classifiers and data points by improving the invariance of perturbations to translation \cite{dong2019evading}. This improves the performance of the attack in the {\it black-box} scenario where the parameters of the target model are not known to the adversary. Forth, smoothness enhances plausible deniability and allows the attacker to disguise APs as natural phenomena such as shadows. In this way, the magnitude of APs can be increased notably since imperceptibility is less important. 

We formulate the problem of constructing smooth APs according to a general definition of smoothness and exploit the geometry of decision boundaries to find computationally efficient solutions.
Our main contributions are the followings:
\begin{itemize}
    \item We propose SmoothFool, a geometry inspired framework for computing smooth APs which exploits the topology of decision boundaries to find efficient APs.
    \item We analyze various properties of smooth APs and validate their effectiveness for both the white-box and black-box attack scenarios. 
    
    \item We show the susceptibility of two major group of defenses against smooth APs by breaking several state-of-the-art defenses.
    
    \item We integrate SmoothFool with previous studies on universal APs and demonstrate the existence of smooth universal APs that generalize well across data samples and network architectures.
\end{itemize}

\section{Related Work}
\subsection{Adversarial Attack} 
Despite the highly non-linear nature of DNNs, they have been observed to exhibit linear characteristics around the actual parameters of the model and the input samples \cite{goodfellow2015explaining, fawzi2017robustness, fawzi2018empirical}. In particular, Goodfellow \etal \cite{goodfellow2015explaining} showed that the prediction of DNNs can be changed drastically by translating the input sample toward the gradient of the classification loss. Hence, they proposed the fast gradient sign method (FGSM) as a single step attack incorporating solely the sign of gradients to craft APs. Kurakin \etal \cite{kurakin2016adversarial} improved the performance of FGSM by adopting an iterative procedure called IGSM.
Moosavi \etal \cite{moosavi2016deepfool} proposed DeepFool to find approximately $\ell_p$-minimal APs by iteratively translating input samples toward the linearized approximation of the closest decision boundary. Our methodology builds on DeepFool to find minimal smooth APs.  

Some prior studies have considered smoothness in adversarial attacks. Sharif \etal \cite{sharif2016accessorize} added a total variation (TV) loss to the main objective of the attack to enhance the physical realizability of the resulting perturbations and showed that smoothness of APs improves their effectiveness for the real-world applications. 
Fong \etal \cite{fong2017interpretable} demonstrated that smoothing important regions in the input example can deteriorate the confidence of prediction. They utilized this observation to interpret the decisions of DNNs.
Hosseini \etal \cite{hosseini2018semantic} proposed constructing semantic adversarial examples by randomly shifting Hue and Saturation components of benign samples in the HSV color space. Dong \etal \cite{dong2019evading} demonstrated that robustness of DNNs to slight translations can be exploited to improve the trasferability of adversarial examples. Interestingly, the final perturbations crafted using their approach exhibited low-pass frequency response. However, their methodology is applicable for a limited level of smoothness since the prediction of DNNs is invariant for solely small translations of the input sample.

Fundamentally, our work differs from previous approaches since we seek to find approximately $\ell_2$-minimal APs capable of offering arbitrary levels of smoothness. Also, our main goal is to formulate and compute smooth APs, not to find smooth adversarial examples, since the latter can critically destroy the structure of images.  
\subsection{Defense Methods}
Since the first observation of APs, their noisy structure has been harnessed to find defense strategies. Several studies have incorporated explicit denoising techniques to mitigate the adversarial effect. Liao \etal \cite{Liao2018defense} showed that the distribution of high-level representations in DNNs provides an effective guidance to denoise adversarial examples and proposed the high-level representation guided denoiser (HGD). Training DNNs using adversarial examples, known as \textit{adversarial training} \cite{goodfellow2015explaining, madry2018toward, tramer2018ensemble}, has been shown to provide a relative adversarial robustness. Adversarial training can be considered as an implicit denoising technique which reduces the sensitivity of predictions to slight changes in the input domain. Manifold learning is another implicit denoising defense. A well-known example for this type of defense is {\it MagNet} \cite{meng2017magnet} which deploys autoencoders for mapping input examples onto the manifold of natural examples. Later, we utilize these defenses to evaluate the effectiveness of smooth APs. %analyze the effectiveness of smooth APs against the aforementioned %explicit and implicit 
%denoising-based defenses.

\section{Smooth Adversarial Perturbations}
\subsection{Problem Definition}
Let $f: \mathbb{R}^n \rightarrow \mathbb{R}^m$ be a classifier mapping input sample $\x \in [0,1]^n$ to $m$ classification scores $f_j(\x)$, associated with each class $j\in\{0, \ldots, {m-1} \}$. The class predicted by the network can be computed as:
\begin{equation}
c(\x) =\argmax_j f_j(\x).
\end{equation}
The problem of constructing smooth APs can be formulated as the following optimization problem:
\begin{equation}
\label{eq_op1}
\begin{split}
\argmin_{\r} ||\r||_2+\lambda\Omega(\r) \text{ 
subject to:}~~~~~~~~~~~~~~~~~ \\
1.~~c(\x+\r)\neq c(\x),~~~~~~~~~~~~~~~~~~~~~~~~ \\
2.~~\x+\r \in [0, 1]^n,~~~~~~~~~~~~~~~~~~~~~~~~~\,
\end{split}
\end{equation}
where $\r \in \mathbb{R}^n$ is the AP, $\Omega(.)$ is a measure of roughness, and $\lambda$ is a Lagrangian coefficient controlling the trade-off between roughness and magnitude of the perturbation. Generally, the roughness of perturbations can be defined based on their local variations. Such variations have an explicit interpretation in the frequency domain where the power of each frequency component captures the specific range of variations. Considering this perspective, we use a frequency response function $H$ to formulate the definition of roughness since it can denote how much each frequency component contributes to the intended roughness. For clarity, we substitute $H$ with $H_{hp}$ to highlight the high-frequency nature of roughness and denote $H_{lp}=1-H_{hp}$ as the complementary low-pass filter which defines the equivalent smoothness.  We use the total energy of the high-frequency components of $\r$ as a general measure of roughness, and define $\Omega$ as:
\begin{equation}
\label{eq_powlowpass}
    \Omega(\r, H_{lp}) := \int_{-\infty}^{+\infty}R(\omega)^2(1-H_{lp}(\omega))^2 d\omega,
\end{equation}
where $R$ is the Fourier transformation of perturbation $\r$, and $H_{lp}$ is the frequency response of a given low-pass filter defining the range of acceptable smoothness, and is a free parameter of the definition. 

The perturbation $\r$ in our problem is represented as a set of spatially discrete APs for each pixel location $u\in\{0, \ldots, {n-1} \}$\footnote{Here we assume the input image $\x$ is a 1D signal, and later in the experiments we adopt all formulations for 2D images.}, and $\Omega$ can be conveniently computed in the spatial domain as:
\begin{equation}
    \label{eq_roughness_discrete}
    \Omega(\r;\h) =||\r-\r*\h||^2_2,
\end{equation}
where $*$ denotes convolution, and $\h$ is the discrete approximation of $H_{lp}$ in the spatial domain. In the rest of the paper, our work builds on this definition of roughness (and, equivalently, smoothness) and aims to find APs which are relatively smooth based on any predefined $\h$ compared to perturbations crafted by other contemporary attacks. Due to the non-convex nature of the problem, we exploit the geometric properties of the decision boundary of DNNs to find a relaxed solution for the optimization problem given in Equation \ref{eq_op1}. 
\begin{figure}
    \centering
    \includegraphics[width=120pt]{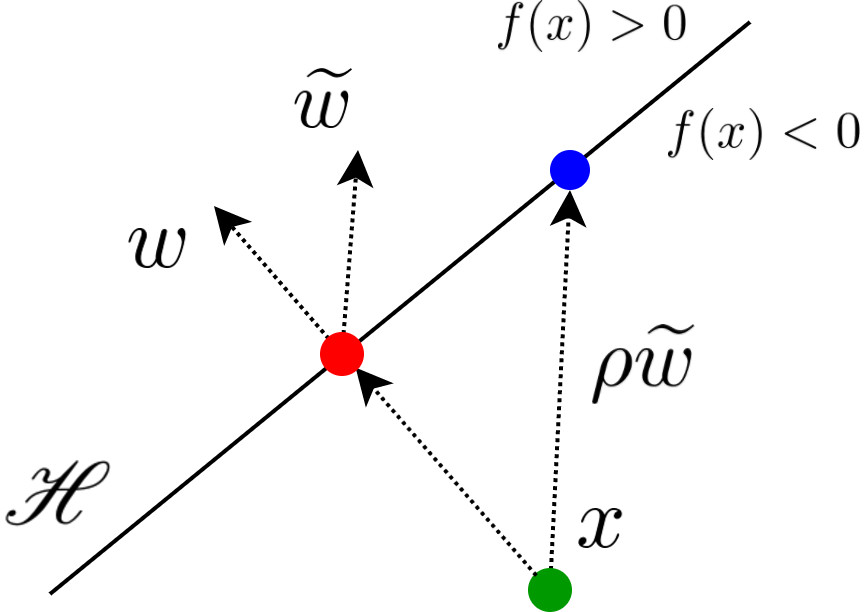}
    \caption{Finding smooth AP for a linear binary classifier. Red and blue dots show the $\ell_2$ projection and smooth projection of $\x$ onto the decision boundary, respectively. For an easier demonstration, $\x$ is assumed to belong to class $-1$.}
    \label{fig:mathfig0}
\end{figure}
\subsection{Linearized solution}

Based on previous findings \cite{moosavi2016deepfool, jetley2018friends, fawzi2017robustness}, the decision boundary of a differentiable classifier, $f$, around $\x$ can be well approximated by a hyperplane passing through the minimal $\ell_2$ adversarial example $\x_p$ corresponding to $\x$, and the normal vector $\w$ orthogonal to the decision boundary at $\x_p$ as $\mathscr{H} \triangleq \{\x: \w^\top(\x-\x_p)=0 \}$. We assume $\x_p$ and, consequently, $\w$ associated with each $\x$ is available, and later we utilize an appropriate contemporary attack to compute $\x_p$ and $\w$. Having $\x_p$ provides two benefits. First, it allows us to linearize the closest decision boundary around $\x$. Second, we can reduce the problem to a binary classification problem, where the goal of the attack would be to compute the smooth perturbation $\r$ which yields $c(\x+\r)=c(\x_p)$. Consequently, we rewrite the optimization problem given in Equation \ref{eq_op1} as:
\begin{equation}
\label{eq_op2}
\begin{split}
\argmin_{\r} ||\r||_2+\lambda\Omega(\r; \h) \text{ 
subject to:}~~~~~~~~~~~~~~~ \\
1.~~\w^\top(\x+\r)-\w^\top\x_p=0,~~~~~~~ \\
2.~~\x+\r \in [0, 1]^n.~~~~~~~~~~~~~~~~~~~~~~~~\,
\end{split}
\end{equation}

In this setup, an efficient solution can be obtained from a smooth projection of $\x$ onto the estimated hyperplane $\mathscr{H}$. Such a projection can be computed by translating $\x$ using the adversarial perturbation $\r = \rho\widetilde{\w}$, where $\widetilde{\w}$ is a smooth approximation of $\w$, and $\rho$ scales $\widetilde{\w}$ to map $\x+\r$ on $\mathscr{H}$ as:
\begin{equation}
\label{eq_rho}
\rho = \dfrac{\w^\top(\x_p-\x)}{\w^\top\widetilde{\w}}.
\end{equation}
Figure \ref{fig:mathfig0} provides a simple visualization of this projection. It worth mentioning that for the linear binary classifier choice of $f$, the optimal smooth perturbation has the closed-form solution: $\r = -\dfrac{f(\x)}{\w^{\top}\widetilde{\w}}\widetilde{\w}$. Generally, the estimation $\widetilde{\w}$ must hold two conditions to provide a valid solution for the linearized problem. First, $\widetilde{\w}$ should not be orthogonal to $\w$. Second, the estimation should remove  high-frequency components of $\w$ in order to keep $\Omega(\rho\widetilde{\w};\h)$ low. Without loss of generality, we consider a low-pass filter $\g$ to estimate $\widetilde{\w}$ by convolution as: $\widetilde{\w} = \g*\w$, since it is easy to compute, and the only condition on $\g$ is that its cut-off frequency should be less than the cut-off frequency of $\h$. 

The final smooth perturbation that can project $\x$ on $\mathscr{H}$ can be computed as:
\begin{equation}
\label{eq_perturbation}
\r = \dfrac{\w^\top(\x_p-\x)}{\w^\top(\g*\w)}(\g*\w).
\end{equation} 
In this formulation, the cut-off frequency of $\g$ is associated with $\lambda$ in the optimization problem given in Equation \ref{eq_op2} since it controls the smoothness of perturbation $\r$. 
\begin{figure}
    \centering
    \includegraphics[width=130pt]{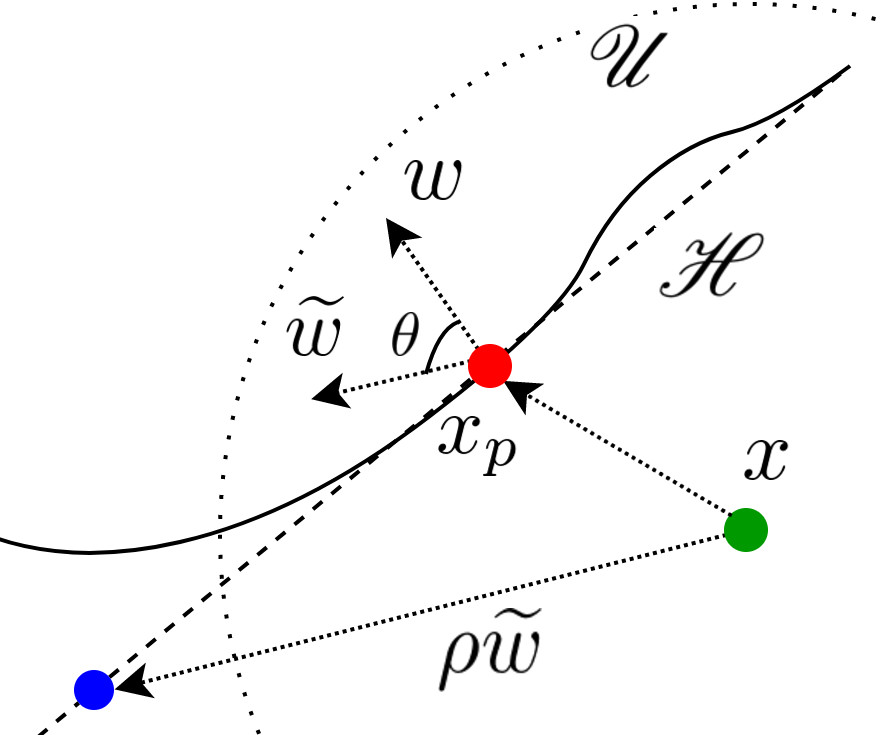}
    \caption{A demonstration of the topology of the decision boundary in the vicinity of data point $\x$. $\mathscr{U}$ illustrates the region where the decision boundary can be assumed to be approximately flat. Smooth projection of $\x$ onto the estimated hyperplane $\mathscr{H}$ often results in a solution out of $\mathscr{U}$.}
    \label{fig:mathfig}
\end{figure}
$\ell_2$-DeepFool \cite{moosavi2016deepfool} constructs adversarial examples which are shown to be a good approximation of the $\ell_2$-minimal adversarial example for an input sample, and the assumption of flat decision boundaries around the constructed examples is believed to be practically valid \cite{moosavi2016deepfool, modas2018sparsefool}. Therefore, we utilize it to generate $\x_p$ and estimate $\w$ using the first order Taylor expansion of $f$ at $\x_p$ as:
\begin{equation}
\label{eq_normalvec}
\w = \nabla f_{c(\x_p)}(\x_p) - \nabla f_{c(\x)}(\x_p).
\end{equation}
In practice, the high-frequency structure of the gradients of DNNs increases the angle between $\w$ and $\widetilde{\w}$. Consequently, $\rho$ in Equation \ref{eq_rho} takes large values which often maps the input sample outside the legitimate range $[0, 1]^n$. In the next section, we propose a smooth clipping technique to overcome this problem.

\begin{algorithm}[t]
\small
\caption{SmoothClip}
\begin{algorithmic}[1]
\State \textbf{input:} Image $\x$, perturbation $\r$, low-pass filter $\g$, step size $\epsilon$.
\State \textbf{output:} Smoothly clipped perturbation $\r_{c}$.
\State Initialize $\r_c \gets \r$.
\While{$\max({\x+\r_c}) > 1$ or $\min({\x+\r_c}) < 0$ }
\State $\m_0 = \mathds{1}_{>0}(-(\x+\r_c))* \g$,
\State $\m_1 = \mathds{1}_{>0}((\x+\r_c)-1)* \g$,
\State $\Delta_1 = \max(\x+\r_c-1)\m_1$, 
\State $\Delta_0 = \min(\x+\r_c)\m_0$, 
\State $\r_c \gets \r_c -\epsilon(\Delta_1+\Delta_0)$,
\EndWhile
\State \textbf{return} $\r_c$.
\end{algorithmic}
\label{alg:Smoothclip}
\end{algorithm}

\begin{figure*}
    \centering
    \includegraphics[width=450pt]{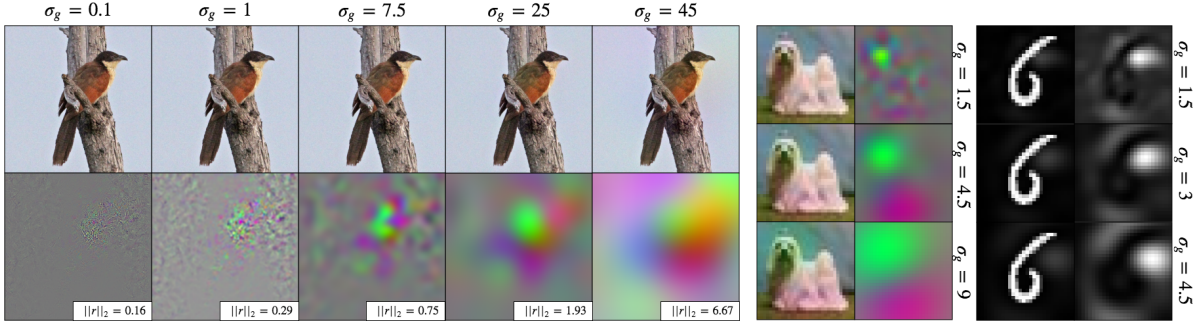}
    \caption{Visual demonstration of increasing smoothness of APs. Each set of images, from left to right, show adversarial examples and smooth APs computed for samples from ImageNet, CIFAR-10, and MNIST datasets on ResNet-101, ResNet-18, and LeNet architectures, respectively. Samples are from \textit{`coucal'}, \textit{`dog'}, and \textit{`6'} classes and misclassified as \textit{`robin'}, \textit{`bird'}, and \textit{`0'}.     
    }
    \label{fig:comparestd}
\end{figure*}

\subsection{Validating Perturbations}
\label{sec:validatingperturbations}
The final adversarial example should reside inside the valid range of the input domain.
% (second condition of the optimization problem in Equation \ref{eq_op2}). 
An ordinary approach to hold this condition, especially in iterative attacks, is to clip the resulting adversarial examples \cite{kurakin2016adversarial, modas2018sparsefool}. The clipping function, $Clip$, takes the constructed adversarial image and truncates each pixel value independently to fall within the valid range of the input space. However, applying this to smooth perturbations as: $\r_c = Clip(\x+\r) - \x$, will deteriorate the smoothness of perturbation. This is because the clipping function truncates each pixel individually and discards the local correlation between neighborhood perturbations. Specifically, this issue happens at edges and high-frequency areas of $\x$ as shown in Figure \ref{fig:clipfig}. A closed-form solution for smooth clipping, which should consider neighborhood correlation of perturbations (based on $\g$), results in a high complexity solution. We propose a simple and iterative approach for smoothly clipping the out-of-bound pixels. In the $i^{th}$ iteration, when the range of $\x + \r^{i}$ remains out of the valid range, we compute masks $\m_0$ and $\m_1$ as indicators of pixels which exceed the valid bound as:
\begin{equation}
\label{eq_ml}
\m_0^{i} = \mathds{1}_{>0}(-(\x+\r^{i})),
\end{equation}
\begin{equation}
\label{eq_mh}
\m_1^{i} = \mathds{1}_{>0}((\x+\r^{i})-1),
\end{equation}
where $\mathds{1}_{>0}(.)$ is an indicator function that outputs $1$ for elements greater than zero. To incorporate the neighborhood correlation of perturbations, we use the exact low-pass filter $\g$ used in Equation \ref{eq_perturbation} to propagate the out-of-bound error to the neighborhood perturbations as: $\m_1^{i} \leftarrow \m_1^{i}* \g$ and $\m_0^{i}\leftarrow\m_0^{i}* \g$. Then, using a step size $\epsilon$ and maximum value of the out-of-bound error, we adjust the perturbation as:
\begin{figure}
    \centering
    \includegraphics[width=180pt]{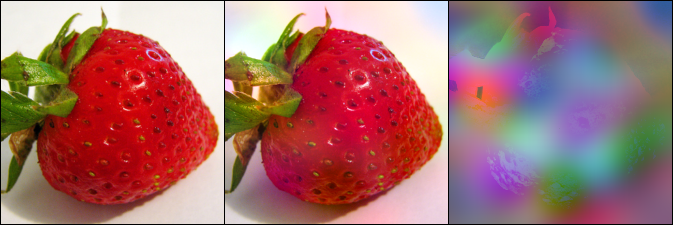}
    \caption{An example of applying a normal clipping on a smooth AP. Left: a benign sample correctly classified as \textit{`strawberry'} by VGG16. Middle: an adversarial example %constructed by the smooth perturbation in Equation \ref{eq_perturbation} 
    classified as \textit{`pineapple'}. Right: the perturbation after normal clipping.}
    \label{fig:clipfig}
    \vspace{-20pt}
\end{figure}
\begin{equation}
\label{eq_ml}
\begin{split}
\r^{i+1} = \r^{i}-\epsilon\max(\x+\r^{i}-1)\m_1^{i}\,\\-\epsilon\min(\x+\r^{i})\m_0^{i}.~~~~~~~~~~
\end{split}
\end{equation}
This iterative algorithm terminates when all pixels in $\x+\r^{i}$ reside within the valid range. We refer to this algorithm as \textit{SmoothClip}, and Algorithm \ref{alg:Smoothclip} summarizes its functionality. 

\subsection{General Solution}
In a general case for a non-linear classifier $f$, there is no guarantee that perturbations computed by Equation \ref{eq_perturbation} cause input samples to pass the actual non-linear boundary. 
Figure \ref{fig:mathfig} demonstrates a visualization of this fact. To overcome this problem, we adopt an iterative procedure. In each iteration, using the closest adversarial example, $\x_p^{i}$, corresponding to the sample $\x^{i}$, we linearize the decision boundary and compute the smooth projection of $\x^{i}$ on the approximated hyperplane using Equation \ref{eq_perturbation}. 
 
Afterward, we smoothly rectify the resulting perturbation, $\r$,
and repeat this procedure until $c(\x^{i})\neq c(\x^{0})$, as detailed in Algorithm \ref{alg:SmoothFool}. Here, the smoothness of the final perturbation depends on the smoothness in each iteration. Consider $r^{tot} = x^i-x^0 = \sum_{j=0}^{i} r^j$, where $i$ is the total number of iterations, and $r^j$ is the $j^{th}$ smooth AP. It can be shown that the roughness of the overall perturbation is bounded as: $\Omega(r^{tot};h)\leq i^2 \max_j\Omega(r^j; h)$. To compute an AP with the desired level of roughness defined by $h$, we select $g$ such that $\forall j: \Omega(r^j; h)\ll ||r^j||^2_2$, \ie,~the cut-off frequency of $g$ should be smaller than $h$. In practice, $\max_j \Omega(r^j; h)\ll i^{-2}$, \ie, even for a significantly smooth choices of $\g$ the algorithm converges in few iterations.

\begin{algorithm}[t]
\small
\caption{SmoothFool}
\begin{algorithmic}[1]
\State \textbf{input:} Image $\x$, low-pass filter $\g$.
\State \textbf{output:} Smooth perturbation $\r$.
\State Initialize $\x^{0} \gets \x$, $i \gets 0$.
\While{$c_f(\x^{0})= c_f(\x^{i})$}
\State $\r_p =$ \text{DeepFool}$(\x^{i})$,
\State $\x_p=\x^{i}+\r_p$,
\State $\w^{i}=\nabla f_{c(\x_p)}(\x_p)-\nabla f_{c(\x)}(\x_p)$,
\State $\widetilde{\w}^{i} = \g * \w^{i}$,
\State $\r^{i}= \dfrac{{\w^{i}}^{\top}(\x_p^{i}-\x^{i})}{{\w^{i}}^{\top}\widetilde{\w}^{i}}\widetilde{\w}^{i}$,
% \State $\r^{i} \gets \alpha\r^{i}$
\State $\r^{i} \gets$\text{SmoothClip}$(\x^{i}, \r^{i}, \g)$,
\State $\x^{i+1} \gets \x^{i} + \r^{i}$, 
\State $i \gets i+1$,
\EndWhile
\State \textbf{return} $\x^{i}-\x^{0}$.
\end{algorithmic}
\label{alg:SmoothFool}
\end{algorithm}

\section{Experiments}
\subsection{Setup}
\label{sec:setup}
We evaluate the performance of SmoothFool on three datasets including the test set of MNIST \cite{lecun2010mnist}, the test set of CIFAR-10 \cite{krizhevsky2009learning}, and $10,000$ samples from the validation set of ILSVRC2012 \cite{deng2009imagenet} (10 images per each class). For the MNIST dataset, a two-layer fully-connected network (FC2) and a LeNet \cite{lecun1998gradient} architecture are used. For the CIFAR-10 dataset, we use a VGG-F \cite{chatfield2014return} and ResNet-18 \cite{he2016deep} architectures. For the ImageNet dataset, we consider VGG16 and ResNet-101. The accuracy of each model on benign samples is shown in Table \ref{tab:generalcomparison}. We set the step size $\epsilon$ of the SmoothClip to $1$, $0.5$, $0.1$ for MNIST, CIFAR-10, and ImageNet respectively, which results in a fast and reasonable performance. 
\begin{figure*}
    \centering
    \includegraphics[width=490pt]{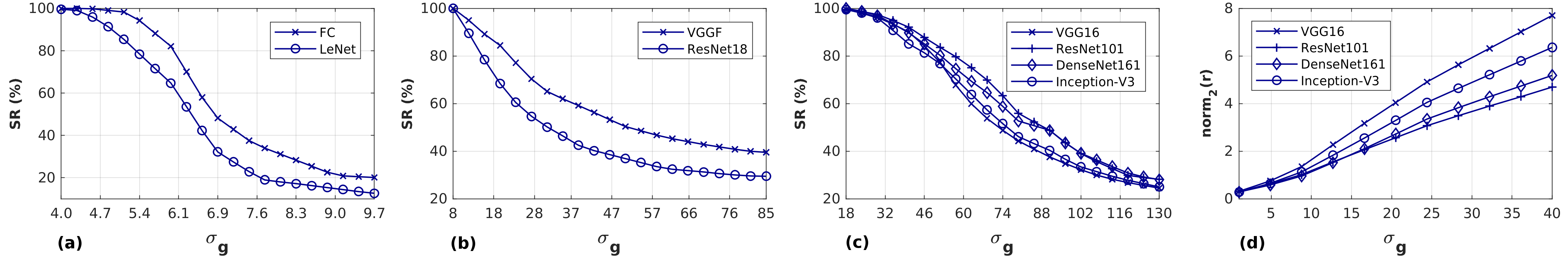}
    \caption{a, b, c) Fooling rate of the attack versus smoothing factor $\sigma_{\g}$ on MNIST, CIFAR-10 and ImageNet, respectively. d) Magnitude of perturbations vs. $\sigma_{\g}$ on ImageNet. }
    \label{fig:srvssigma}
    \vspace{-10pt}
\end{figure*}

\noindent{\bf Defining smoothness.} We define smoothness based on the Gaussian blur function since it is practical and the cut-off frequency can be easily changed by modifying the standard deviation.  %\textcolor{red}{In Sub-section ??? we consider two other choices of smoothing functions for an ablation study on the type of smoothing.}
We assume $\h$ and $\g$ to be Gaussian blur filters with isotropic standard deviations $\sigma_{\h}$ and $\sigma_{\g}$, respectively. In this setup, selecting any ${\sigma_{\g}\!>\!\sigma_{\h}}$ will minimize the roughness defined by $\h$. Increasing $\sigma_{\g}$ improves the smoothness of APs but reduces the performance of the attack. To implement the Gaussian kernel, we set the kernel width to $5\sigma$. % and increases the computation time of the algorithm.
%We consider similar smoothing functions as employed by Dong \etal \cite{dong2019evading} for our evaluations including a uniform, linear, and Gaussian functions. The free parameter which defines the smoothness of the functions for the uniform and linear cases are the kernel size, and for the Gaussian function is the standard deviation, $\sigma$. For the Gaussian function, we set the kernel size to be $5\sigma$. %Therefore, we assume $\h$ and $\g$ to be Gaussian blur filters with isotropic standard deviations $\sigma_{\h}$ and $\sigma_{\g}$, respectively.
%For the uniform and linear cases, selecting 
%In this setup, selecting any ${\sigma_{\g}\!>\!\sigma_{\h}}$ will minimize the roughness defined by $\h$. Increasing $\sigma_{\g}$ results in smoother APs, but theoretically reduces the performance of the attack and increases the computation time of the algorithm.

\begin{table*}[]
\centering
\footnotesize
\begin{tabular}{clccc|cccc|cccc|cccc}
\hline
\multirow{2}{*}{\textbf{Dataset}} &
\multirow{2}{*}{\textbf{Network}} &
\multirow{2}{*}{\begin{tabular}{@{}c@{}}\textbf{Acc.} \\ (\%)\end{tabular}} &
\multicolumn{2}{c|}{\textbf{F.Rate} (\%)} & \multicolumn{4}{c|}{$\boldsymbol{\overbar{\Omega}\times10^3$ @ $\sigma_h\!=\!1}$} & \multicolumn{4}{c|}{$\boldsymbol{\overbar{\Omega}_n\times10^3$ @ $\sigma_h\!=\!1}$} & \multicolumn{4}{c}{$\boldsymbol{\mathbb{E}_{\x}[{||\r_{\x}||_2}]$ @ $\sigma_{\g}}$}  \\
 &&& \textbf{IS} \rule{0pt}{8pt} & \textbf{CS} & \textbf{DF} & \textbf{IS} & \textbf{CS} & \textbf{SF} & \textbf{DF} & \textbf{IS} & \textbf{CS} & \textbf{SF} & \textbf{DF} & \textbf{IS} & \textbf{CS} & \textbf{SF}  \\ \hhline{=================}
\multirow{2}{*}{MNIST} & 1-FC2 & $98.6$ & $98.0$ & - & $783$ & $591$ & - & $\textbf{334}$  & $511$ & $308$ & - & $\textbf{63}$ & $1.17$ & $2.81$ & - & $1.76$ \\
 & 2-LeNet & $99.1$ & $94.1$ & - & $890$ & $677$ & - & $\textbf{352}$  & $532$ & $338$ & - & $\textbf{68}$ & $1.23$ & $3.18$ & - & $2.32$ \\\hline
\multirow{2}{*}{CIFAR-10} & 3-VGG-F & $93.1$ & $85.4$ & $93.4$ & $288$ & $203$ & $184$ & $\textbf{114}$ & $891$ & $216$ & $163$ & $\textbf{57}$ & $0.19$ & $3.14$ & $4.42$ & $1.60$ \\ 
& 4-ResNet-18 & $93.3$ & $87.8$ & $89.1$ & $310$ & $206$ & $199$  & $\textbf{127}$ & $959$ & $287$ & $175$ & $\textbf{65}$ & $0.21$ & $3.63$ & $4.93$ & $1.60$ \\\hline
\multirow{2}{*}{ImageNet} & 5-VGG16 & $71.5$ & $57.8$ & $91.1$ & $111$ & $89$ & $104$ & $\textbf{29}$ & $871$ & $358$ & $238$ & $\textbf{51}$ & $0.25$ & $4.90$ & $7.71$ & $0.58$ \\ 
 & 6-ResNet-101 & $77.3$ & $62.6$ & $92.7$ & $108$ & $83$ & $95$ & $\textbf{16}$ & $827$ & $300$ & $207$ & $\textbf{36}$ & $0.28$ & $4.82$ & $7.56$ & $0.55$\\ 
 & 7-ResNet-152 & $78.3$ & $60.1$ & $91.9$ & $116$ & $91$ & $88$ & $\textbf{20}$ & $845$ & $391$ & $229$ & $\textbf{38}$ & $0.29$ & $5.12$ & $8.16$ & $0.57$ \\ 
 & 8-DenseNet161 & $77.6$ & $66.0$ &  $90.3$& $107$ & $79$ & $86$ & $\textbf{16}$ & $795$ & $309$ & $212$ & $\textbf{35}$ & $0.28$ & $3.47$ & $5.58$ & $0.57$ \\ 
 & 9-InceptionV3 & $77.4$ & $65.5$ & $90.6$ & $149$ & $85$ & $92$ & $\textbf{12}$ & $658$ & $293$ & $223$ & $\textbf{48}$ & $0.32$ & $4.50$ & $8.10$ & $0.35$  \\ 
\hline 
\end{tabular}
\caption{Comparing SmoothFool (SF) to DeepFool (DF) \cite{moosavi2016deepfool}, iterative smooth 
(IS) \cite{sharif2016accessorize, eykholt2018robust}, and color-shift (CS) \cite{hosseini2018semantic} attacks. To satisfy smoothness based on $\sigma_{\h}\!=\!1$, $\sigma_{\g}$ is set to $2$ for all datasets. Fooling rates of SF and DF are $>\!99.9\%$ on all datasets. %Magnitude of perturbations constructed by CS are computed in the HSV color space. 
}.
\label{tab:generalcomparison}
\end{table*} 
%%%%%%%%%%%%%%%%%%%%%%%%%%
\noindent{\bf Comparisons.}
For $\sigma_{\g}\!\ll\!1$, the proposed approach converges to DeepFool   \cite{moosavi2016deepfool}. Hence, we use it as a baseline to compare magnitude of the generated perturbations.
For the second baseline, we develop an attack based on \cite{sharif2016accessorize} by replacing the classical TV loss term with the roughness penalty $\Omega(\r;\h)$ from Equation \ref{eq_roughness_discrete} to provide a fair comparison framework as:
\begin{equation}
\label{eq_comparisonattack}
\argmin_{\r}  \big(-J_c(f(\x+\r), y_{\x})+\lambda_{s}\Omega(\r;\sigma_{\h})\big),
\end{equation}
where $J_c$ is the cross-entropy loss function, and $y_{\x}$ denotes the ground truth label of sample $\x$. We refer to this method as the \textit{iterative smooth} (IS) attack, and optimize it using gradient descent with a initial step size (learning rate) of $10^{-3}$, and decay of $0.5$ per each $100$ iterations. We set $\lambda_{s}$ to $0.1$, $0.01$ and $0.05$ for MNIST, CIFAR-10, and ImageNet, respectively, which results in the most possible smooth perturbations for $\sigma_{\h}\!=\!1$. We consider this attack as the second baseline. We also compare the proposed method to the semantic adversarial examples given in  \cite{hosseini2018semantic}, and refer to it as the \textit{color-shift} (CS) attack, and consider it as the third baseline. We set the number of random trails of the CS algorithm to 100. Since CS adds perturbations in the HSV color space, we compute the average magnitude of perturbations for this attack in the HSV space to provide a fair comparison (the magnitude of APs in RGB color space is observed to be approximately 10 times greater).

\noindent{\bf Evaluation metrics.}
%To compare results, 
We measure the fooling rate of the attack and average smoothness of constructed APs. The fooling rate is defined on the set of correctly classified benign samples since it provides a more robust measure to evaluate the attack. For the sake of brevity of explaining results, we define $\sigma_{\g}^{A\%}$ as the maximum value of $\sigma_{\g}$ (minimum among network architectures) that results in a $A \%$ fooling rate. In order to evaluate the smoothness of constructed APs, we measure the expected roughness $\overbar{\Omega}=\mathbb{E}_{\mathscr{D}_s}[\Omega(\r_{\x},\h)]$, 
where $\r_x$ is the AP constructed for $\x$, and $\mathscr{D}_s$ is the set of successfully attacked samples. This measure is sensitive to the magnitude of perturbations. Thus, we develop a second measure by normalizing $\Omega$ over the total power of perturbations as: $\overbar{\Omega}_n\!=\!\mathbb{E}_{\mathscr{D}_s}[\Omega(\r_{\x},\h)/||\r_{\x}||^2_2]$.
Indeed, $\overbar{\Omega}_n$ relatively measures how much of the total power of the APs is occupied by high-frequency components according to $\h$.

\subsection{General Performance} 

Figure \ref{fig:srvssigma} (a-c) shows the fooling rates of SmoothFool versus $\sigma_{\g}$. We observe that the pair $(\sigma_{\g}^{100\%}, \sigma_{\g}^{20\%})$ for MNIST, CIFAR-10 and ImageNet is $(4.1, 7.8)$, $(8.4, 124.4)$ and $(19.3, 165.3)$, respectively. As expected, the fooling rate is highly dependent on the smoothing factor $\sigma_{\g}$. However, the fooling rate remains high for significantly large (compared to the size of the input image) values of $\sigma_{\g}$ on ImageNet and CIFAR-10. For instance, $\sigma_{\g}^{50\%}$ for
CIFAR-10 is $32.8$ which is approximately equal to the width of input images and shows that it is possible to fool the classifier on $50\%$ of samples solely by adding a carefully selected constant value to all pixels of each color channel.
\begin{figure}
    \centering
    \includegraphics[width=210pt]{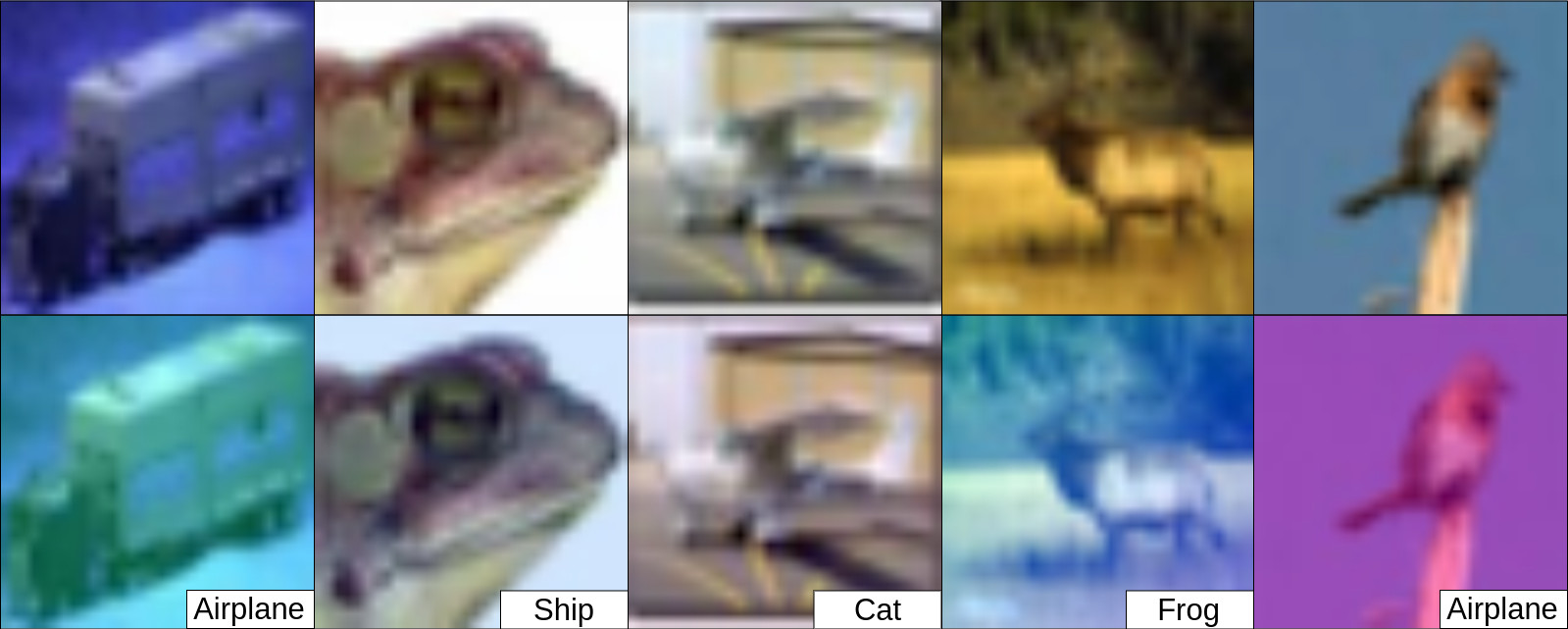}
    \caption{Examples of extremely smooth adversarial perturbations computed for ResNet-18 and CIFAR-10 dataset with $\sigma_{\g}=200$. %For large values of $\sigma_{\g}$ the resulting AP becomes a constant value all over the image showing similar behavior in RGB which were shown for HSV in \cite{hosseini2018semantic}.
    }
    \label{fig:cifarlargesigmasamples}
\end{figure}
The magnitude of smooth APs versus smoothness is depicted in Figure \ref{fig:srvssigma} (d). Increasing smoothness results in larger magnitudes of APs since the projection of $\widetilde{\w}$ onto $\w$ will become smaller. %Therefore, a perturbation computed in each iteration of the main algorithm will take a greater angle to the optimal perturbation and consequently $\rho$ increases. 
However, smoothness of perturbations allows larger magnitudes since they are not as perceptible when compared to the noisy structure of contemporary APs.  % Contemporary attacks often constrain the magnitude of APs to keep them imperceptible to the human eye. Their noisy structures require lower magnitudes compared to smooth adversarial perturbations.     
\begin{table}[]
\centering
\footnotesize
\begin{tabular}{ccccccc}
% & \multicolumn{6}{c}{\textbf{Fooling rate (\%)}} \\ \cline{2-7}
 & \multicolumn{3}{c}{\textbf{VGG-F}\rule{0pt}{8pt}} & \multicolumn{3}{c}{\textbf{ResNet-18}} \\ \hline
\multicolumn{1}{c|}{\multirow{2}{*}{\textbf{Class}}} & \multicolumn{3}{c|}{$\sigma_{\g}$} & \multicolumn{3}{c}{$\sigma_{\g}$} \\
\multicolumn{1}{c|}{} & $20$ & $60$ & \multicolumn{1}{c|}{$100$} & $20$ & $60$ & $100$ \\ \hhline{=======}
\multicolumn{1}{c|}{airplane} & $75.0$ & $29.5$ & \multicolumn{1}{c|}{$25.0$} & $67.3$ & $34.7$ & $28.2$ \\
\multicolumn{1}{c|}{automobile} & $\textbf{58.7}$ & $\textbf{10.8}$ & \multicolumn{1}{c|}{$\textbf{6.5}$} & $\textbf{33.3}$ & $\textbf{6.6}$ & $\textbf{2.2}$ \\
\multicolumn{1}{c|}{bird} & $93.4$ & $65.2$ & \multicolumn{1}{c|}{$\underline{52.1}$} & $65.1$ & $41.8$ & $32.5$ \\
\multicolumn{1}{c|}{cat} & $\underline{100}$ & $58.3$ & \multicolumn{1}{c|}{$43.7$} & $65.3$ & $21.1$ & $17.3$ \\
\multicolumn{1}{c|}{deer} & $87.2$ & $\underline{60.0}$ & \multicolumn{1}{c|}{$49.0$} & $78.0$ & $40.2$ & $36.0$ \\
\multicolumn{1}{c|}{dog} & $78.7$ & $48.9$ & \multicolumn{1}{c|}{$40.4$} & $75.1$ & $\underline{55.5}$ & $\underline{52.7}$ \\
\multicolumn{1}{c|}{frog} & $88.8$ & $51.1$ & \multicolumn{1}{c|}{$46.6$} & $68.9$ & $44.8$ & $41.3$ \\
\multicolumn{1}{c|}{horse} & $74.5$ & $29.0$ & \multicolumn{1}{c|}{$23.6$} & $70.3$ & $25.9$ & $22.2$ \\
\multicolumn{1}{c|}{ship} & $79.0$ & $32.5$ & \multicolumn{1}{c|}{$25.5$} & $\underline{81.4}$ & $35.1$ & $29.6$ \\
\multicolumn{1}{c|}{truck} & $73.9$ & $32.6$ & \multicolumn{1}{c|}{$21.7$} & $68.5$ & $31.4$ & $22.8$ \\
\multicolumn{1}{c|}{all} & $83.8$ & $45.8$ & \multicolumn{1}{c|}{$37.6$} & $67.2$ & $33.0$ & $27.9$
\\\hline
\end{tabular}
\caption{Per-class fooling rate (\%) of SmoothFool for three values of $\sigma_{\g}$ on the CIFAR-10 dataset. Bold and underlined values show the fooling rate on classes with highest and lowest robustness against smooth APs, respectively.}
\label{tab:perclasssuccessrate}
\end{table}    

We observe in Table \ref{tab:generalcomparison} that SmoothFool with $\sigma_{\g}\!=\!2$ on all datasets, crafts significantly smoother (based on $\overbar{\Omega}$ and $\overbar{\Omega}_n$ with $\sigma_{\h}\!=\!1$) APs compared to the baseline attacks for the smoothness, while the magnitudes of APs are solely $1.8$x larger than the state-of-the-art $\ell_2$-minimal APs crafted by DeepFool. %In addition, the execution time of the algorithm detailed in Table \ref{tab:executiontime} shows that the proposed method computes APs (with fooling rate ${>99\%}$) at least $20$x faster than the IS method (with fooling rate ${\simeq 65\%}$ on ImageNet). The execution time of the algorithm is on the same order as DeepFool, which shows that the algorithm finds solutions in a same number of iterations as DeepFool. However, the execution time of the algorithm increases as it searches for smoother APs as demonstrated in Table \ref{tab:executiontime}. 
%Comparing the fooling rate on different architectures shows that increasing the capacity of a model makes it harder to craft smooth APs which is the same as the observation made before for conventional attacks \cite{madry2018toward}.
Figure \ref{fig:comparestd} shows some examples of smooth APs computed for different levels of smoothness. We observe that each class responds differently as the smoothness of APs increases. Table \ref{tab:perclasssuccessrate} shows the per-class fooling rate of the attack on CIFAR-10. Smooth perturbations at $\sigma_{\g}=100$ fool the VGG-F classifier on more than $50\%$ of samples of the \textit{`bird'} class, while they are approximately not effective for the \textit{`car'} class. This shows that some classes are severely sensitive to smooth perturbations while other exhibit lower sensitivity. %In addition, this suggests that DNNs discriminate input samples based on spatial features of different sizes which are particularly assigned to each class. 
The network architecture has a direct effect on this observation since the most sensitive class to smooth APs for each specific value of $\sigma_{\g}$ is different among network architectures.%different for VGG-F and ResNet-18.

\begin{table}[]
\centering
\footnotesize
\begin{tabular}{cccccccc}
& \multicolumn{6}{c}{\textbf{Fooling rate under defense (\%) }} \\ \cline{2-7}
Defense  & IGSM \rule{0pt}{8pt} & DF & CS & $\text{SF}_1$ & $\text{SF}_2$ & $\text{SF}_3$ \\ \hhline{=======}
 Adv.  & $32.6$   & $15.6$ & $64.5$ & $58.6$ & $70.7$ & $\textbf{78.0}$ \\
 PGD  & $21.0$  & $12.3$ & $61.4$ & $57.2$ & $67.3$ & $\textbf{72.8}$ \\
 Ens. &  $18.7$  & $14.0$ & $62.2$ & $54.5$ & $62.8$ & $\textbf{73.6}$ \\
 SAT & $22.8$ & $37.2$ & $21.0$ & $11.5$ & $42.9$ & $\textbf{53.4}$ \\
 HGD  &  $9.3$ & $11.2$ & $46.9$ & $43.7$ & $57.2$ & $\textbf{66.2}$ \\
 MagNet &  $10.7$ & $8.9$ & $25.1$ & $46.4$ & $\textbf{65.5}$ & $52.6$ \\ \hline
\end{tabular}
\caption{Evaluating attacks under different defense strategies on a ResNet-18 trained on CIFAR-10. $\text{SF}_1$, $\text{SF}_2$, and $\text{SF}_3$ denote the proposed algorithm with $\sigma_{\g}$ of $1$, $3$, and $5$, respectively.}
\label{tab:perfunderdefense}
\end{table}

Figure \ref{fig:cifarlargesigmasamples} demonstrates some examples of extremely smooth APs on CIFAR-10, showing a similar behavior (in RGB color space) as color-shifted adversarial examples \cite{hosseini2018semantic}. However, as the method in \cite{hosseini2018semantic} randomly shifts Hue and Saturation of benign samples, it often generates odd adversarial examples such as \textit{`blue apples'} or \textit{`red lemons'} which are no longer adversarial examples since the conceptual evidence of objects is destroyed. However, since SmoothFool finds relatively small smooth perturbations, the whole concept of an object will not change drastically after the attack.
\paragraph{Performance under white-box defenses.}

Here, we evaluate the effectiveness of smooth perturbations against defense methods. First, we evaluate the attack under defenses based on adversarial training on FGSM~(Adv.) \cite{goodfellow2015explaining}, projectile gradient descent (PGD)~\cite{madry2018toward} and ensemble (Ens.)~\cite{tramer2018ensemble} adversarial examples. 
We consider an additional defense of training on adversarial examples computed by the proposed SmoothFool with $\sigma_{\g}=1$, and refer to it as Smooth Adversarial Training (SAT). 
Second, we consider the high-level guided denoiser (HGD)~\cite{Liao2018defense} as a denoising based defense and MagNet~\cite{meng2017magnet} as a defense which evaluates adversarial examples using a learned distribution of natural samples. %We also test the performance under MagNet~\cite{meng2017magnet} defense to explore the closeness of crafted ad. 
In all experiments, we assume that attacks have zero knowledge about the defense models.

Table \ref{tab:perfunderdefense} shows the performance of SmoothFool under defenses. Results suggest that increasing the smoothness of APs elevates the chance of bypassing defense methods. Such a characteristic had been observed before in adversarial examples constructed by spatial transformations \cite{xiao2018spatially, dabouei2018fast}. Smooth APs with $\sigma_{\g}=5$ successfully bypass HGD defense and defenses based on adversarial training on more than $60\%$ of samples. Similarly, the CS attack shows significant robustness against all defenses except MagNet. A reasonable explanation is that although the CS attack generates relatively smooth APs compared to conventional attacks, changing the Hue and Saturation of images considerably pushes samples outside the distribution of natural samples leaned by MagNet. SmoothFool bypasses MagNet by a notable margin which indicates the closeness of generated samples to the distribution of natural images. However, for large values of $\sigma_{\g}$, the magnitude of smooth APs takes large values, and thus, degrades the fooling rate of SmoothFool against MagNet defense. Furthermore, we observe that SAT defense provides a relative robustness against smooth APs constructed by $\sigma_{\g}=1$, but is susceptible to smoother perturbations. This suggest that the frequency components of APs can play a crucial role in bypassing adversarial training defenses trained on examples constructed by APs of different frequency components.     
\begin{figure} 
    \centering
    \includegraphics[width=230pt]{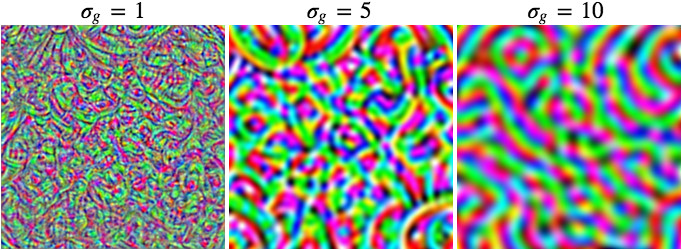}
    \caption{Smooth universal APs crafted for VGG16 architecture (best viewed in color). }
    \label{fig:my_label}
\end{figure}

\noindent{\bf Black-box performance and ablation on smoothing functions.} Here, we evaluate the black-box performance of smooth APs. Since our algorithm computes $\ell_2$-minimal perturbations, we scale smooth APs to have the maximum $\ell_{\infty}$-norm of $16$ for pixel values in range $255$ based on the conventional setting for black-box attacks on ImageNet \cite{dong2019evading}. We consider two additional smooth functions including linear and uniform kernels to evaluate the effect of smoothing functions on fooling rates and transferability of APs. The uniform kernel of size $k$ has all values equal to $\tfrac{1}{k^2}$. The linear kernel of size $k$ has the maximum value of $\tfrac{4}{k^2}$ at the center and minimum value of zero at edges. Any other value is the linear interpolation of the maximum and minimum values. 

Table \ref{tab:blackbox} presents the results for this experiment. The fooling rate of attacks is $100\%$ when the source and target models are the same. This suggests that the type of smoothing functions does not constrain the performance of APs. Hence, a broad range of smoothing functions can be deployed for generating smooth APs. Transferability of adversarial examples consistently improves as the smoothness of perturbations increases. This demonstrates that smoothness increases the transferability of adversarial examples for black-box attacks which validates the results reported by Dong \etal \cite{dong2019evading}

\begin{table}[]
    \centering
    \scriptsize

    \begin{tabular}{c|c|c|ccc}
Net. & Smoothing  & Param.  & VGG16 & ResNet101 & Inc-V3 \\\hhline{======}
& - & - & $100$ & $12.6$ & $8.9$ \\\cline{2-6}

\multirow{6}{*}{\rotatebox[origin=c]{90}{VGG16}} & \multirow{2}{*}{Gaussian} & $\sigma = 5$ & $100$  & $15.8$ & $13.6$  \\ 
 &  & $\sigma =10$ & $100$ & $20.7$ & $15.3$ \\\cline{2-6}
 & \multirow{2}{*}{Linear} & $k=25$ & $100$ & $17.7$ & $11.6$  \\
 &  & $k=50$ & $100$ & $23.5$ & $14.1$ \\\cline{2-6}
 & \multirow{2}{*}{Uniform} & $k=25$ & $100$ & $16.8$ & $12.0$ \\
 &  & $k=50$ & $100$ & $21.8$ & $14.6$ \\ \hline
 & - & - & $15.2$  & $100$ & $15.0$  \\\cline{2-6}
\multirow{6}{*}{\rotatebox[origin=c]{90}{ResNet101}} & \multirow{2}{*}{Gaussian} & $\sigma = 5$ & $17.0$ & $100$ & $18.8$  \\
 &  & $\sigma = 10$ & $19.9$  &  $100$ & $22.5$\\\cline{2-6}
 & \multirow{2}{*}{Linear} & $k=25$ & $19.8$ & $100$ & $16.9$ \\
 &  & $k=50$ & $22.8$ & $100$ & $19.7$ \\\cline{2-6}
 & \multirow{2}{*}{Uniform} & $k=25$ & $18.6$ & $100$ & $17.3$ \\
 &  & $k=50$ & $21.0$ & $100$ & $22.1$\\\hline
\end{tabular}

    \caption{Transferability of smooth perturbations for black-box attack. The size of Gaussian kernels is $k=5\sigma$. Columns show source networks and attack parameters, and rows show the target models.    }
    \label{tab:blackbox}
\end{table}

\noindent{\bf Universal adversarial perturbations.}

We integrate the proposed approach with the universal adversarial perturbations (UAP) \cite{moosavi2017universal} to explore the possibility of finding smooth UAPs. The implementation detail and the integrated algorithm is available in the Supplementary. We compute smooth UAPs for VGG16 and then evaluate thier transferability on four other networks including ResNet-101, ResNet-151, DenseNet-161, and Inception-V3. 

Table \ref{tab:uap} demonstrates the performance of smooth UAPs versus smoothness. % it is possible to craft smooth universal APs capable of transferring across data points and network architectures. 
Increasing smoothness enhances the transferability of APs across both the data points and network architectures. The transferability on 3 networks deteriorates for $\sigma_g>5$. We attribute this observation to the theoretical fact that increasing smoothness also increases the magnitude of APs. Hence, with the same threshold for the maximum $\ell_{\infty}$-norm of smooth UAPs, there always exist a $\sigma_g$ after which the transferability decreases.

\begin{table}[]
\centering
\footnotesize
\begin{tabular}{cccccc} 
  $\sigma_{\g}$ &  VGG16 \rule{0pt}{8pt} & RNet101 & RNet152 & DNet161 & Inc-V3 \\ \hhline{======}
$0$ & $78.3$ & $64.8$  & $63.4$ & $52.9$ & $54.6$ \\
$1$ & $79.6$ & $66.0$ & $66.8$ & $53.2$ & $57.8$ \\
$5$ & $82.2$ & $\boldsymbol{69.9}$ & $\boldsymbol{70.3}$ & $\boldsymbol{57.6}$ & $58.6$ \\
 $10$ & $\boldsymbol{84.5}$ & $68.7$ & $69.1$ & $55.9$ & $\boldsymbol{61.6}$ \\\hline %\hline

\end{tabular}
\caption{Transferability of universal smooth APs computed for VGG16 accross data points and network architectures.}
\label{tab:uap}
\end{table}
\section{Conclusion}
In this study, we explored the vulnerability extent of DNNs to smooth adversarial perturbations by proposing SmoothFool, a framework for computing $\ell_2$-minimal smooth APs. The methodology is developed based on a broad definition of smoothness and can be extended to pose any frequency-domain constraint on perturbations. Through extensive experiments, we demonstrated that smooth adversarial perturbations are robust against two major group of defense strategies. Smoothness also improves the transferability of adversarial examples across network architectures and data points. Furthermore, our results suggest that class categories exhibit variable susceptibility to smooth perturbations which can help interpret the decision of DNNs for different categories.
{\small
\bibliographystyle{ieee}
\bibliography{egbib}
}

\clearpage

\section{Setup for smooth UAPs.}
We use the setup in \cite{moosavi2017universal} and perform two modifications to its main algorithm to find smooth UAPs. First, we replace DeepFool \cite{moosavi2016deepfool} with the SmoothFool algorithm. Second, we replace the clipping method, based on projection operator, with SmoothClip. SmoothClip algorithm smoothly clips $\r$ to make sure $\x+\r$ resides in $[0, 1]$. However, for the smooth UAP $\v$, SmoothClip should clip $\v$ to [$-\xi, \xi$]. Since the new bound is not dependent on $\x$ anymore, we replace $\x$ with an all-zero matrix, $\textbf{\underline{0}}$, of a same size as $\x$. Furthermore, we normalize $\v$ before the clipping and denormalize it after the clipping. This allows us to utilize SmoothClip without modifying its algorithm. Smooth perturbation in each iteration is projected smoothly to the $l_\infty$-ball of $\xi=10/255$ around the original sample. The $\epsilon$ of SmoothClip is set to $0.1$, and the universal fooling rate $\tau$ is set to $0.9$. For computing smooth UAP, we exploit $m=10,000$ benign samples from the validation set of ILSVRC2012 \cite{deng2009imagenet}. Algorithm \ref{alg1} details the algorithm for finding smooth UAPs.

\section{Further results}
\begin{figure}
\centering
    \includegraphics[width=230pt]{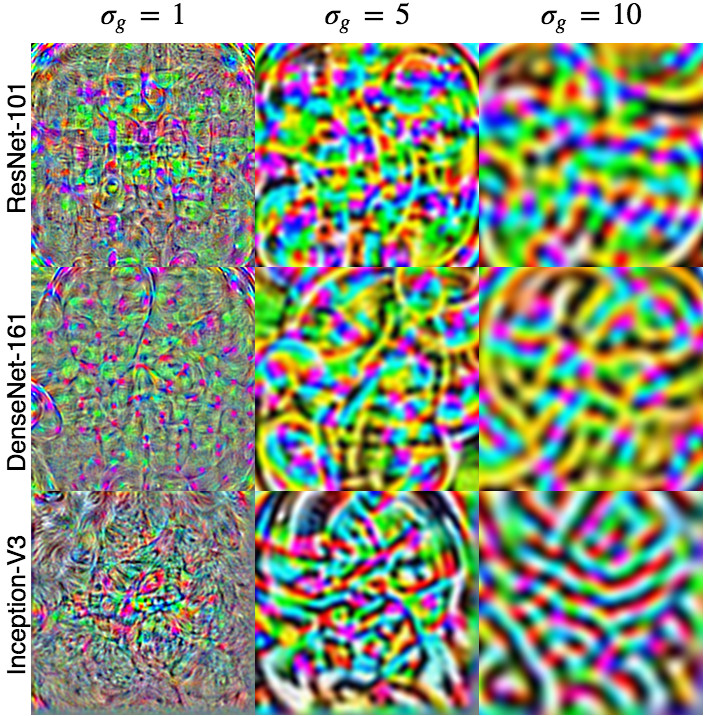}
    \caption{Smooth UAPs computed for three network architectures on ImageNet. Rows and columns show the results on different smoothing levels and network architectures, respectively. 
    }
        \label{fig:uaps}
\end{figure}

\begin{figure*}
    \includegraphics[width=480pt]{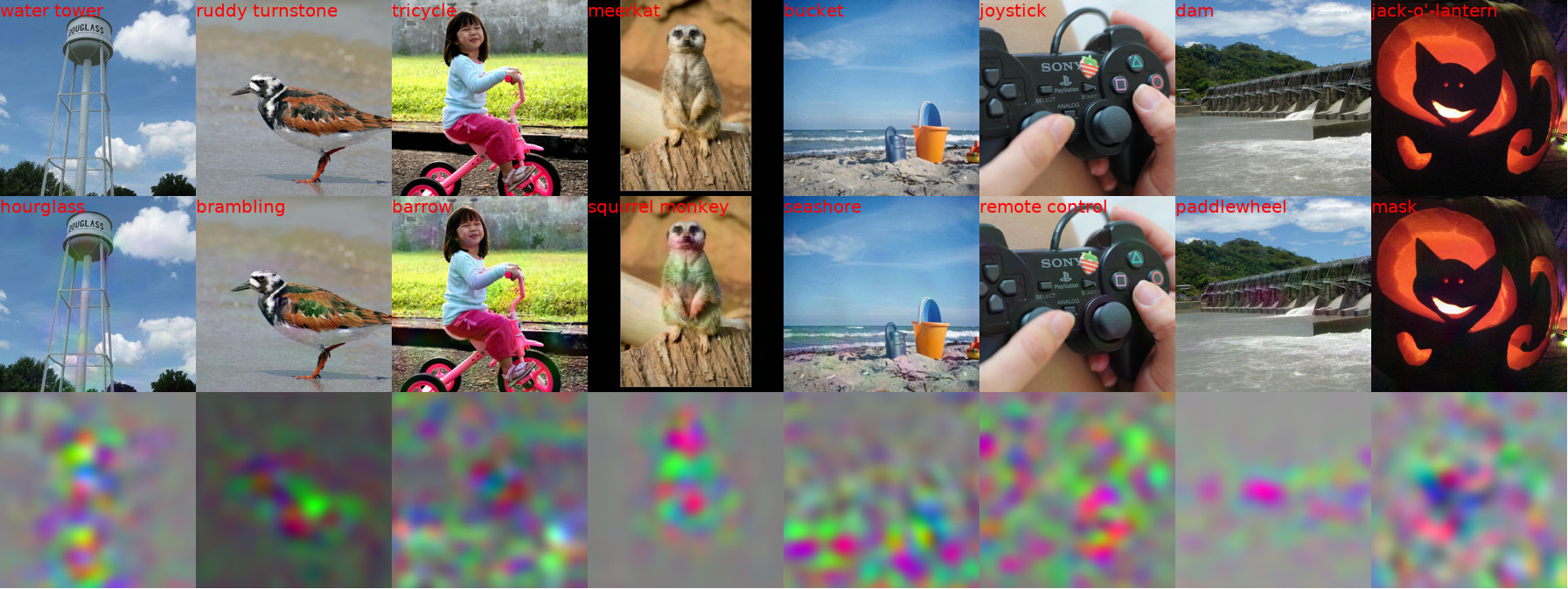}
    \caption{Network architecture: \textbf{VGG-16}, $\sigma_{\g}=\textbf{10}.$
    }
    \label{fig1}
\end{figure*}

\begin{figure*}
    \includegraphics[width=480pt]{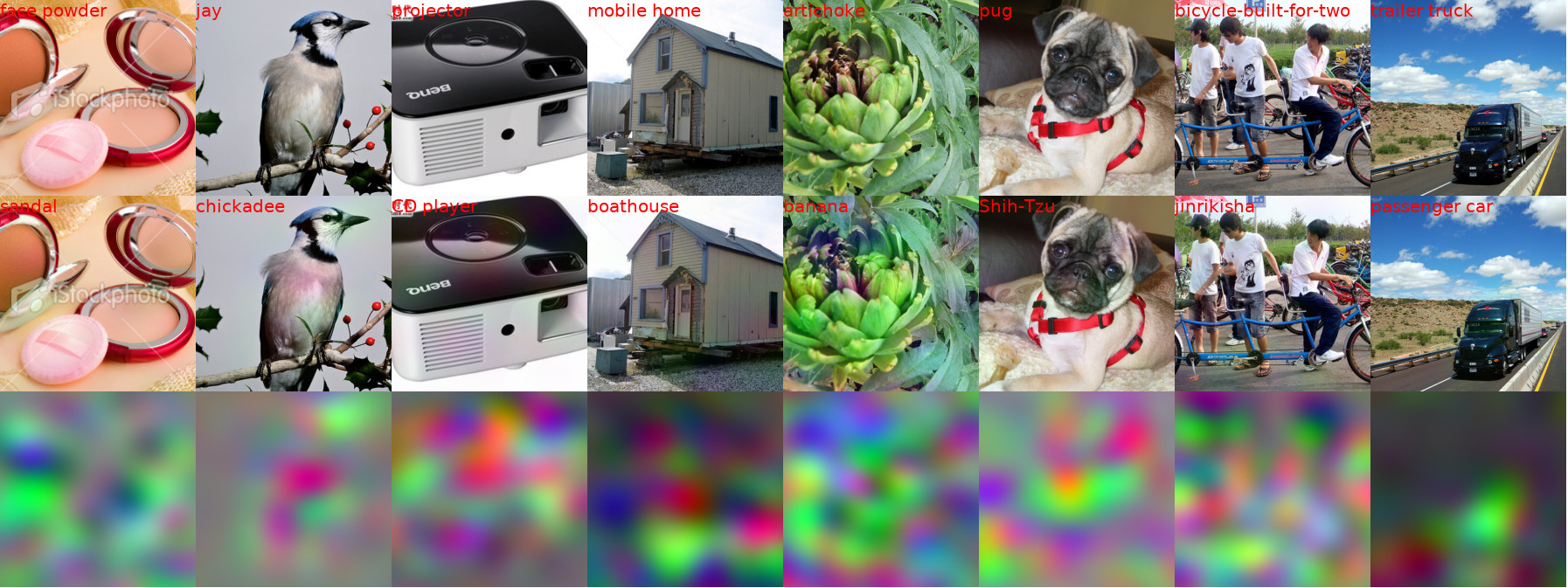}
    \caption{Network architecture: \textbf{VGG-16}, $\sigma_{\g}=\textbf{20}.$
    }
        \label{fig2}
\end{figure*}

\begin{figure*}
    \includegraphics[width=480pt]{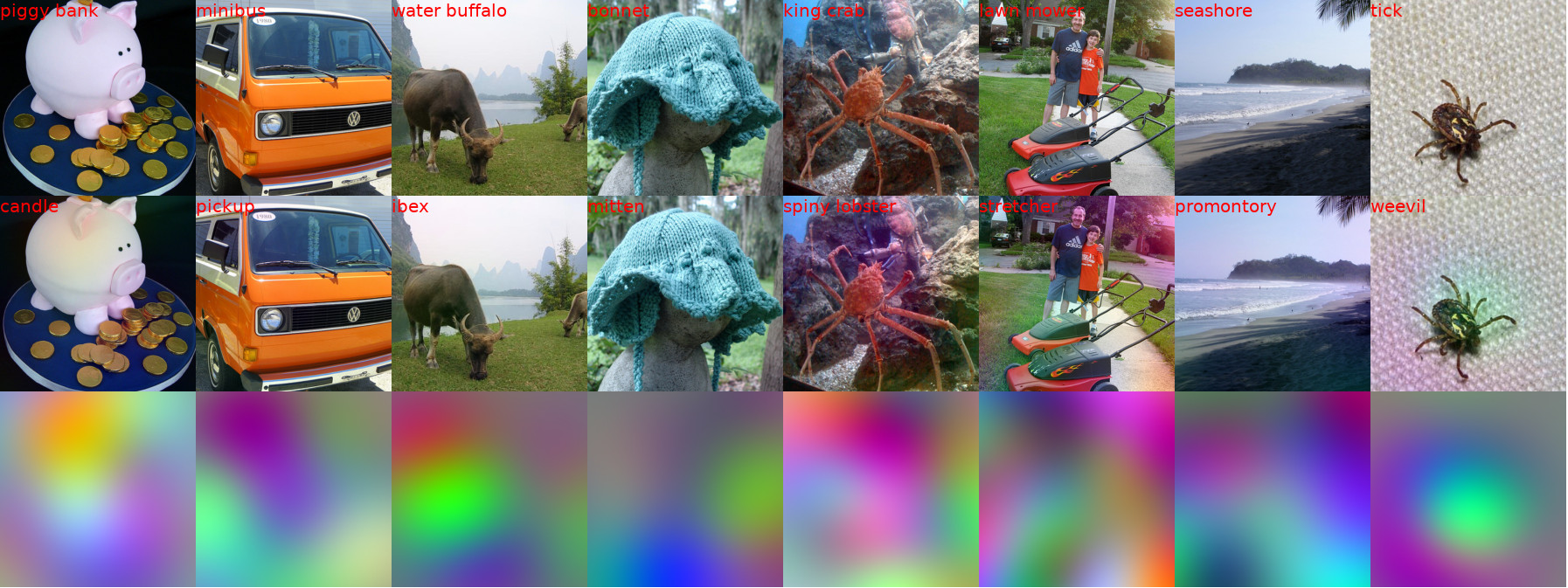}
    \caption{Network architecture: \textbf{VGG-16}, $\sigma_{\g}=\textbf{50}.$
    }
        \label{fig3}
\end{figure*}

\begin{figure*}
    \includegraphics[width=480pt]{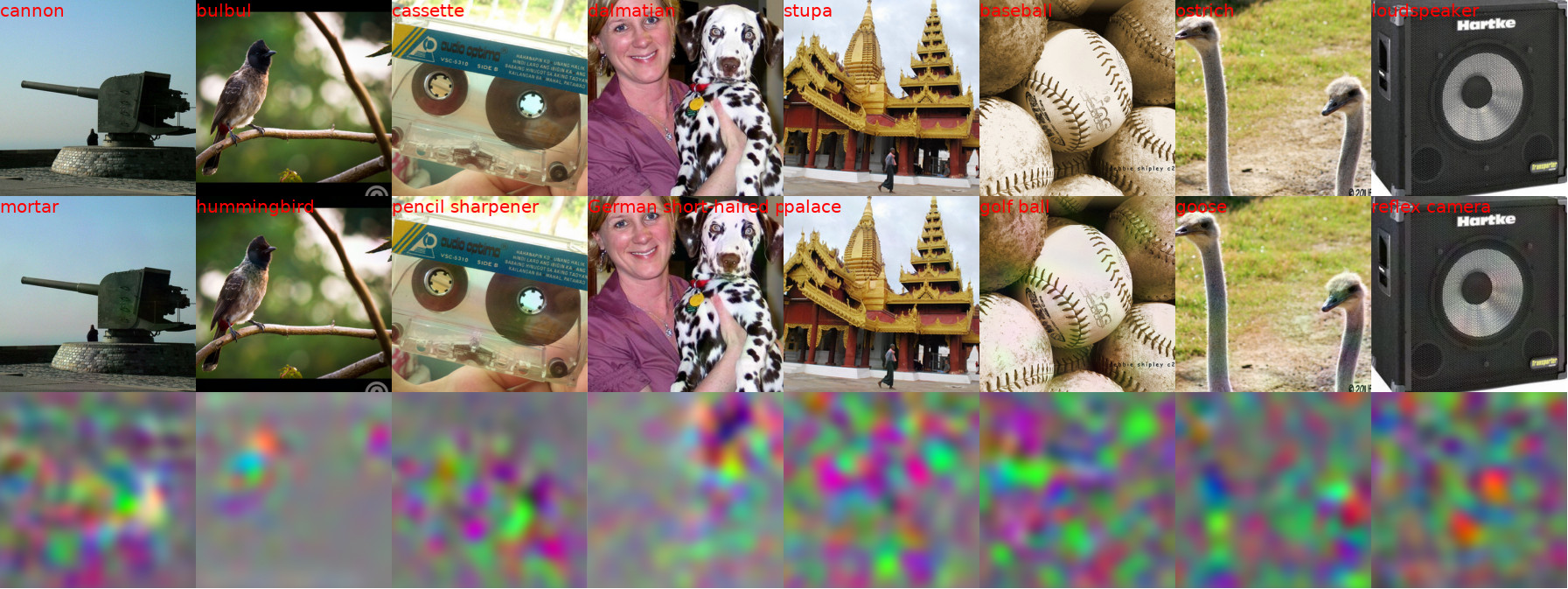}
    \caption{Network architecture: \textbf{ResNet-101}, $\sigma_{\g}=\textbf{10}.$
    }
        \label{fig4}
\end{figure*}

\begin{figure*}
    \includegraphics[width=480pt]{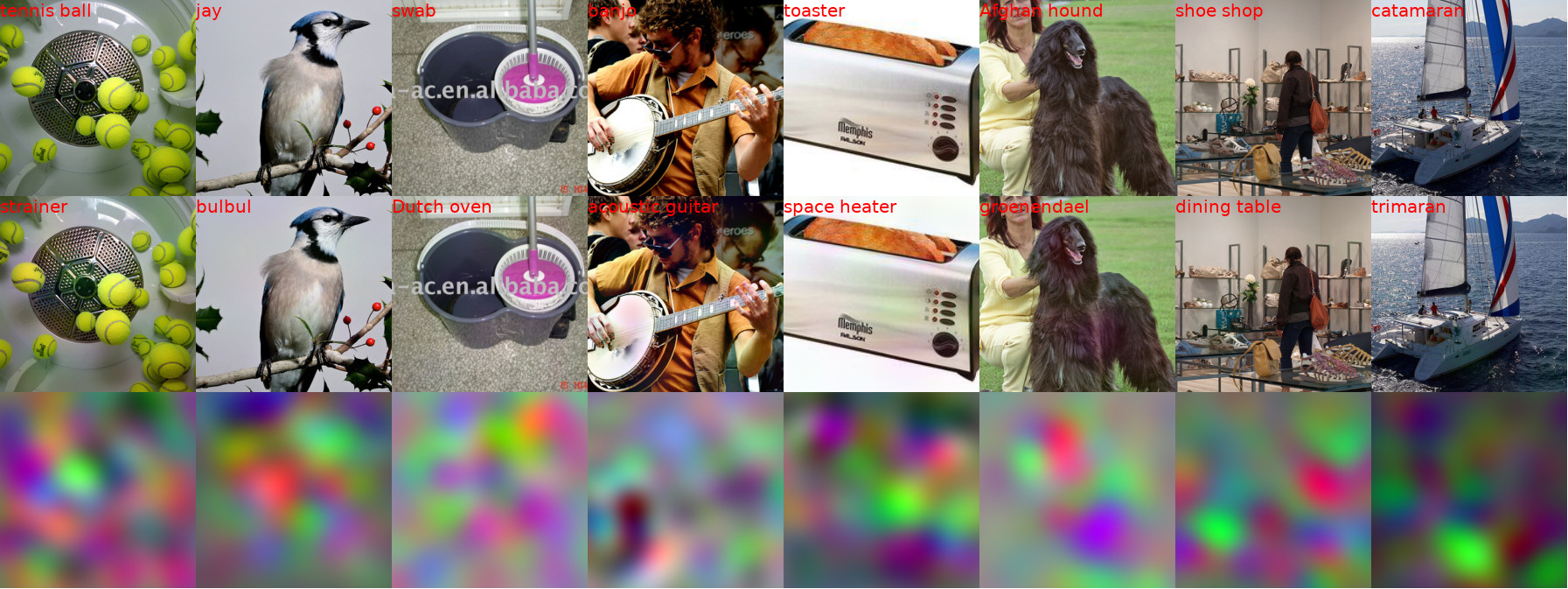}
    \caption{Network architecture: \textbf{ResNet-101}, $\sigma_{\g}=\textbf{20}.$
    }
        \label{fig5}
\end{figure*}

\begin{figure*}
    \includegraphics[width=480pt]{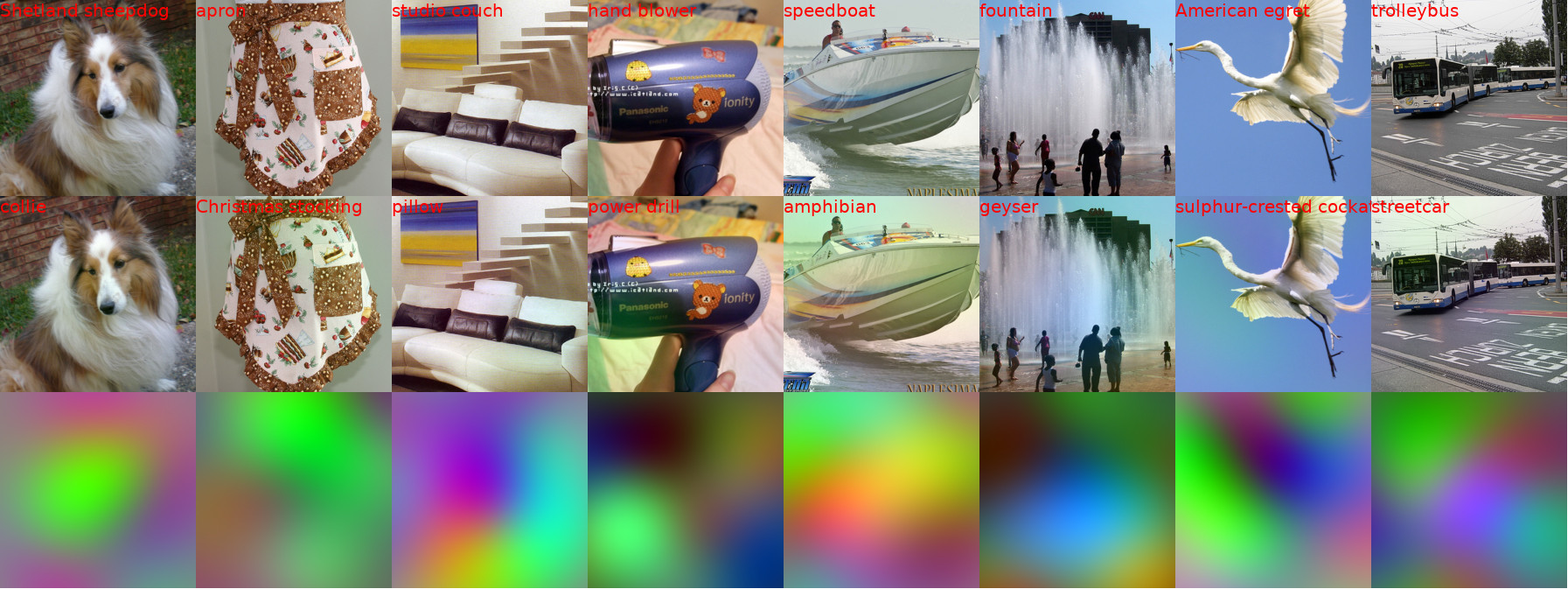}
    \caption{Network architecture: \textbf{ResNet-101}, $\sigma_{\g}=\textbf{50}.$
    }
        \label{fig6}
\end{figure*}

\begin{figure*}
    \includegraphics[width=480pt]{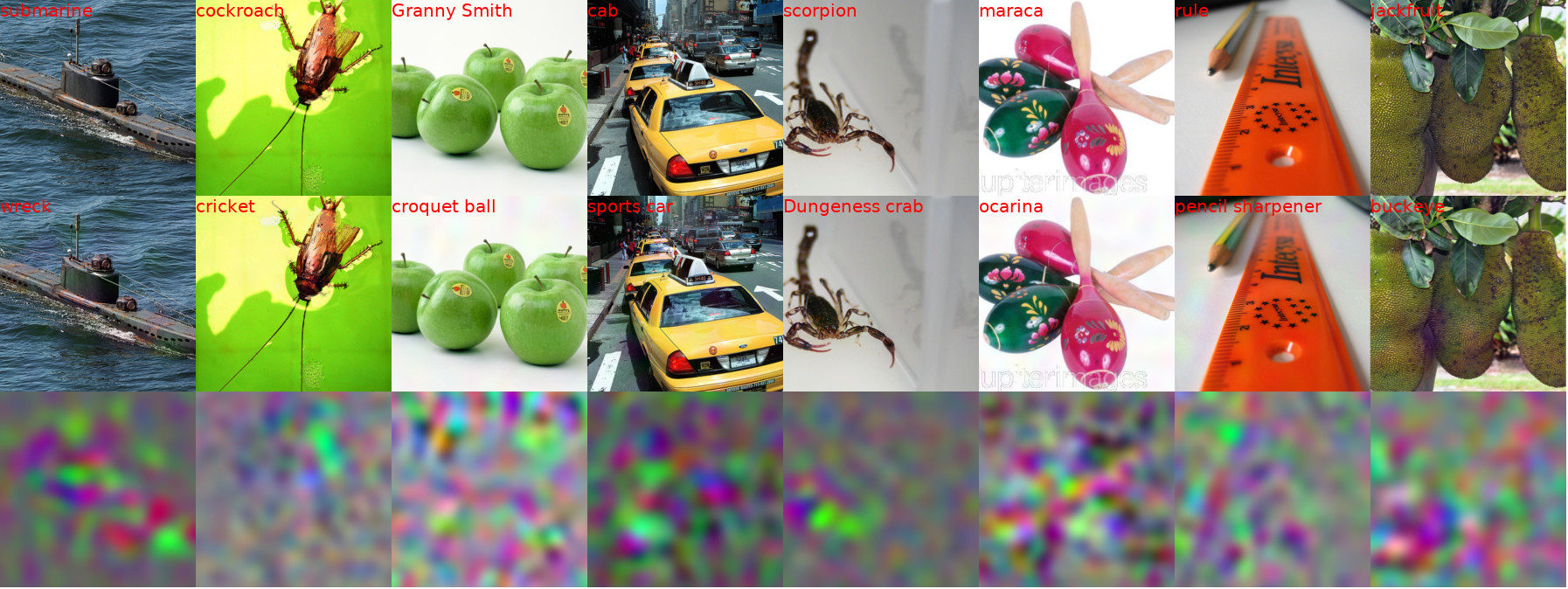}
    \caption{Network architecture: \textbf{DenseNet-161}, $\sigma_{\g}=\textbf{10}.$
    }
        \label{fig7}
\end{figure*}

\begin{figure*}
    \includegraphics[width=480pt]{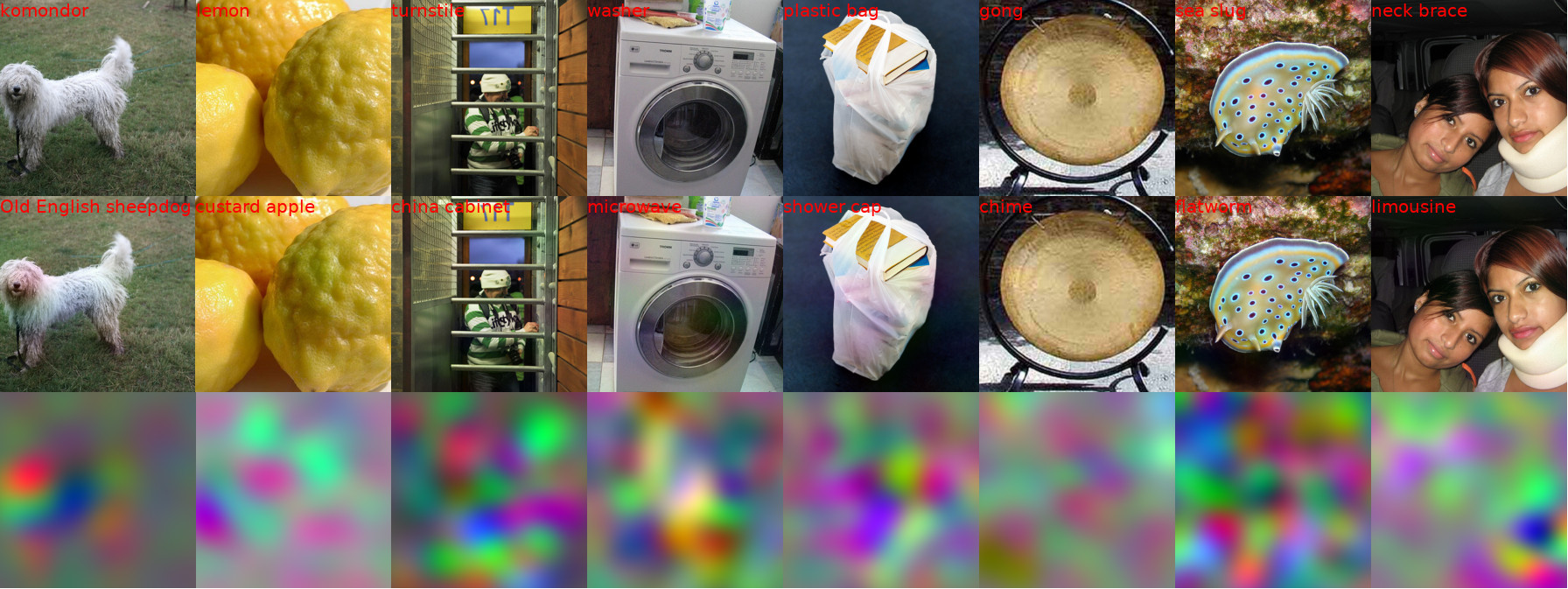}
    \caption{Network architecture: \textbf{DenseNet-161}, $\sigma_{\g}=\textbf{20}.$
    }
        \label{fig8}
\end{figure*}

\begin{figure*}
    \includegraphics[width=480pt]{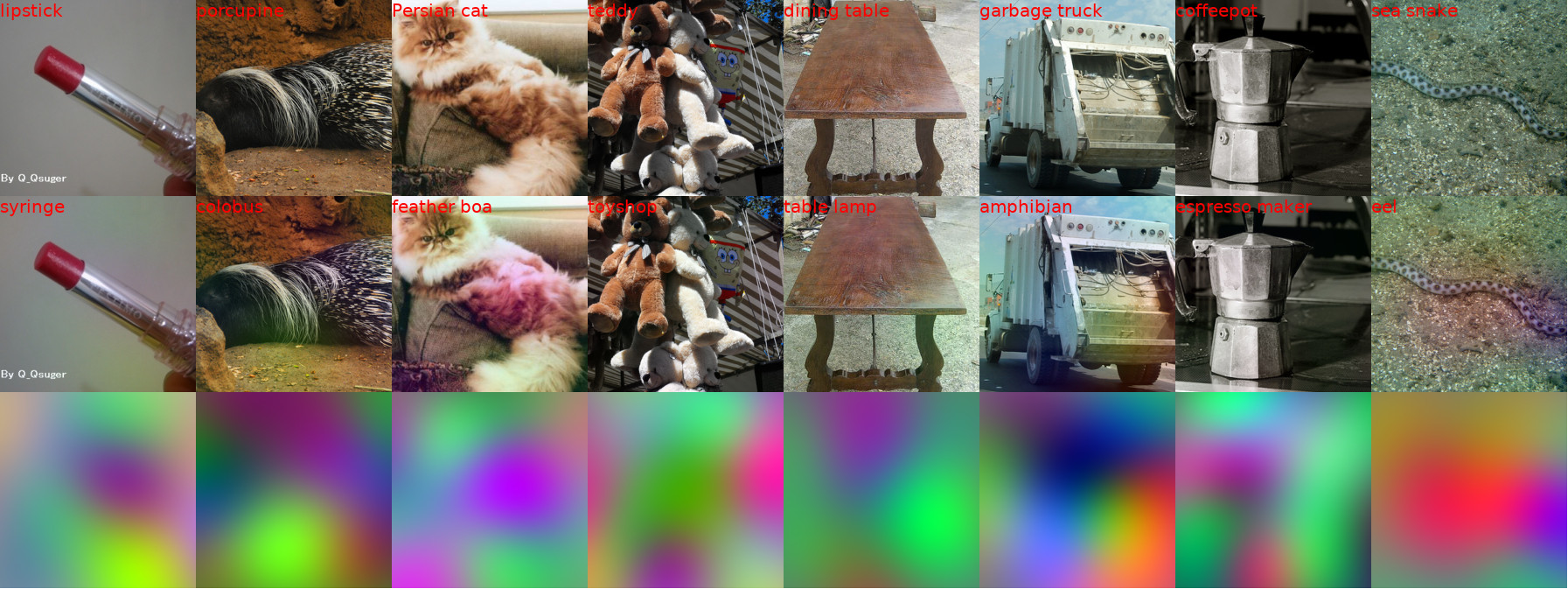}
    \caption{Network architecture: \textbf{DenseNet-161}, $\sigma_{\g}=\textbf{50}.$
    }
        \label{fig9}
\end{figure*}

\begin{figure*}
    \includegraphics[width=480pt]{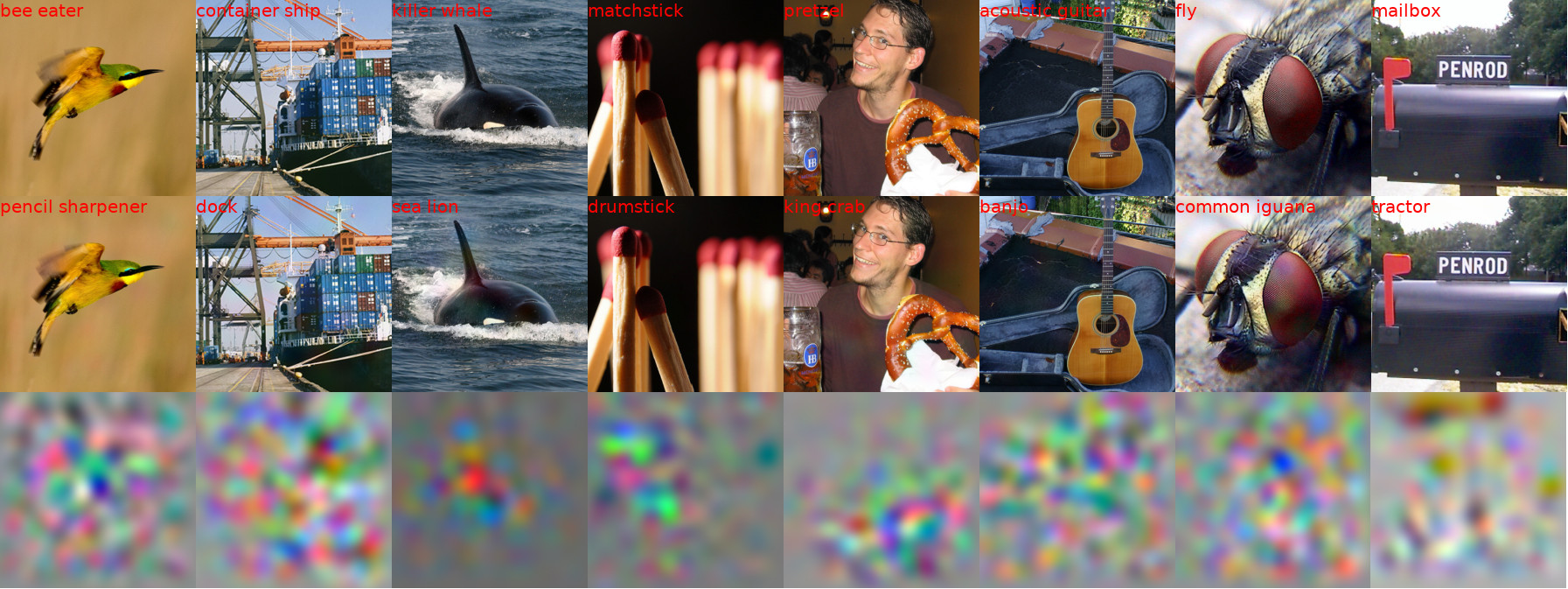}
    \caption{Network architecture: \textbf{Inception-V3}, $\sigma_{\g}=\textbf{10}.$
    }
        \label{fig10}
\end{figure*}

\begin{figure*}
    \includegraphics[width=480pt]{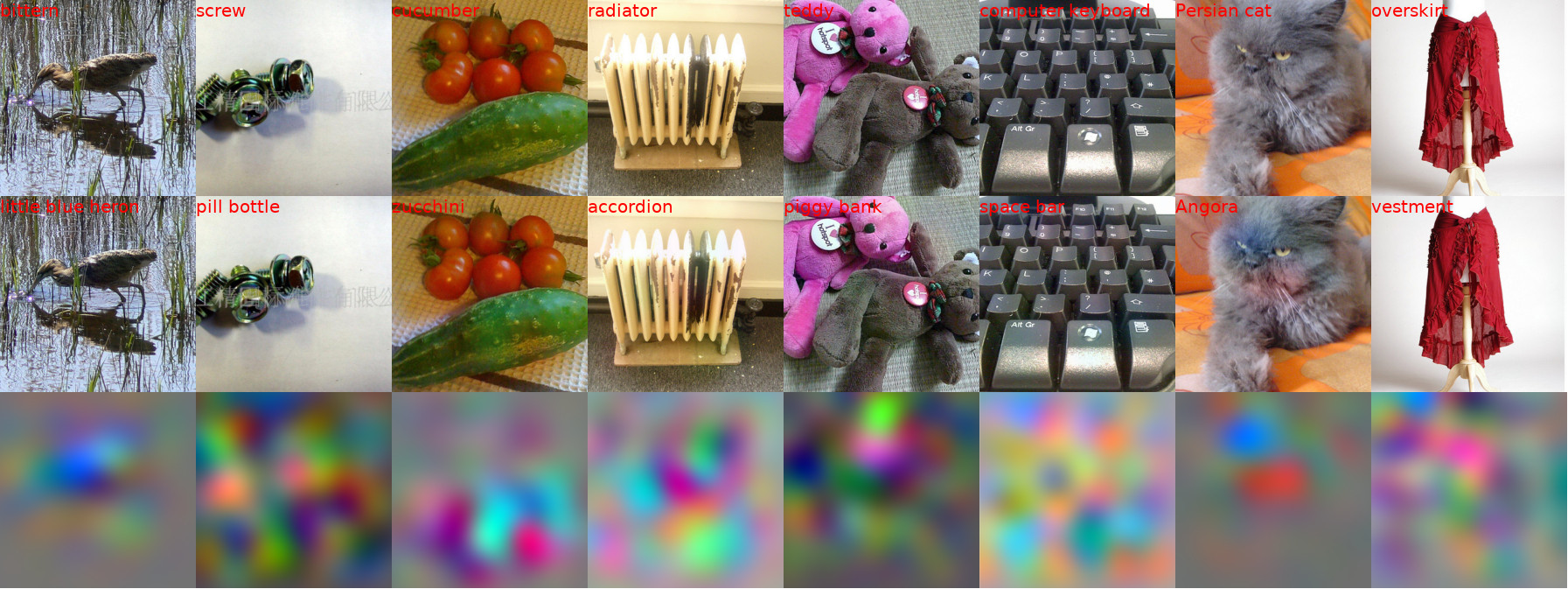}
    \caption{Network architecture: \textbf{Inception-V3}, $\sigma_{\g}=\textbf{20}.$
    }
        \label{fig11}
\end{figure*}

\begin{figure*}
    \includegraphics[width=480pt]{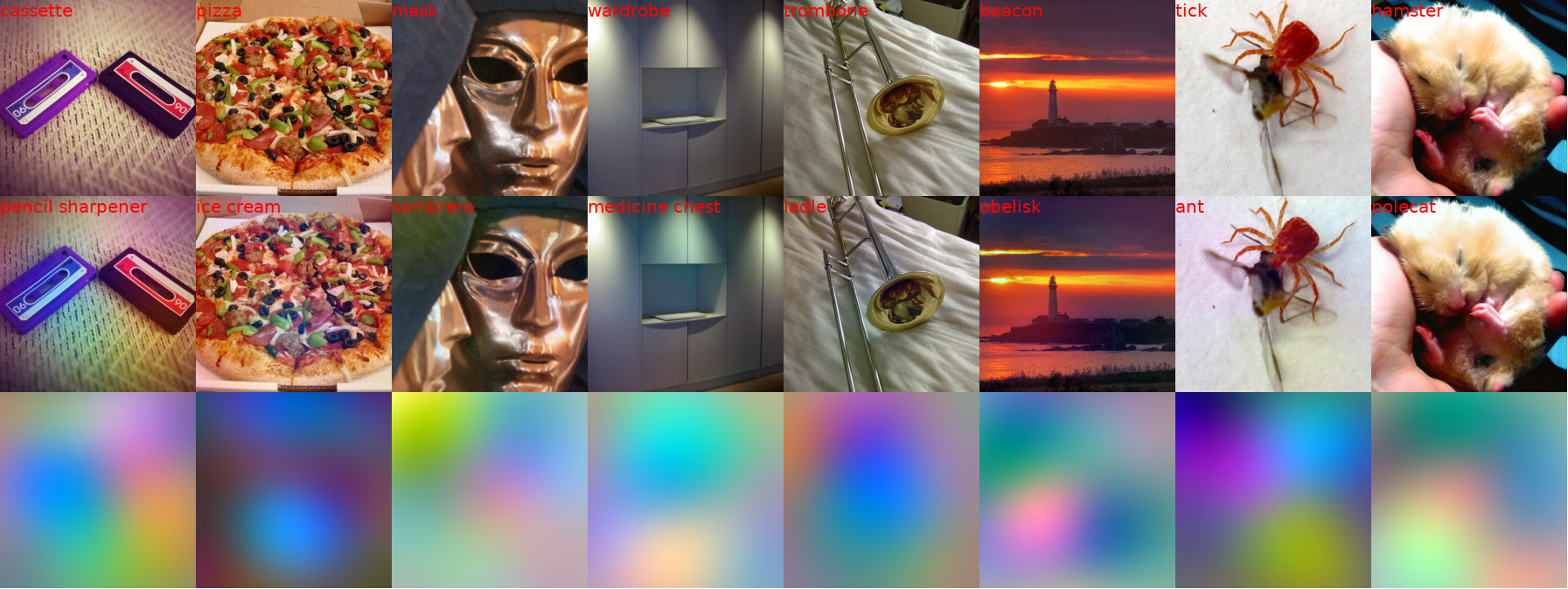}
    \caption{Network architecture: \textbf{Inception-V3}, $\sigma_{\g}=\textbf{50}.$
    }
        \label{fig12}
\end{figure*}

\begin{figure*}
    \includegraphics[width=480pt]{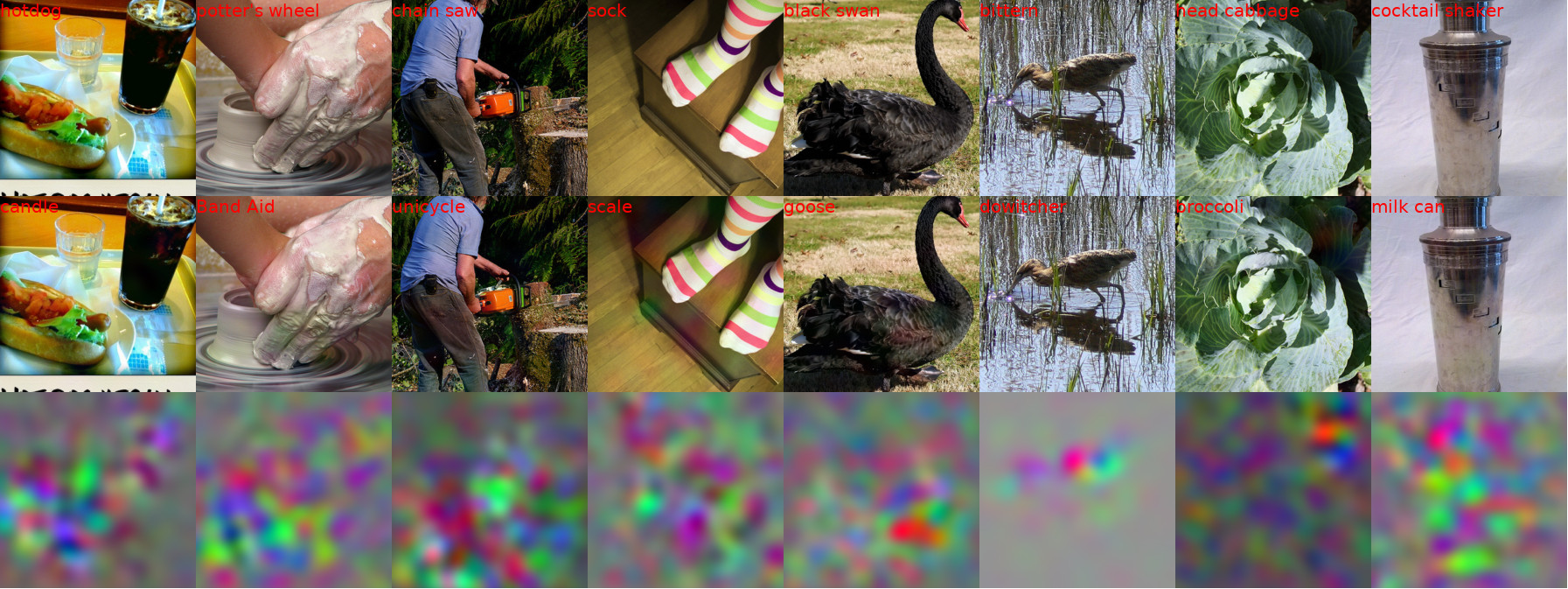}
    \caption{Network architecture: \textbf{ResNet-152}, $\sigma_{\g}=\textbf{10}.$
    }
        \label{fig13}
\end{figure*}

\begin{figure*}
    \includegraphics[width=480pt]{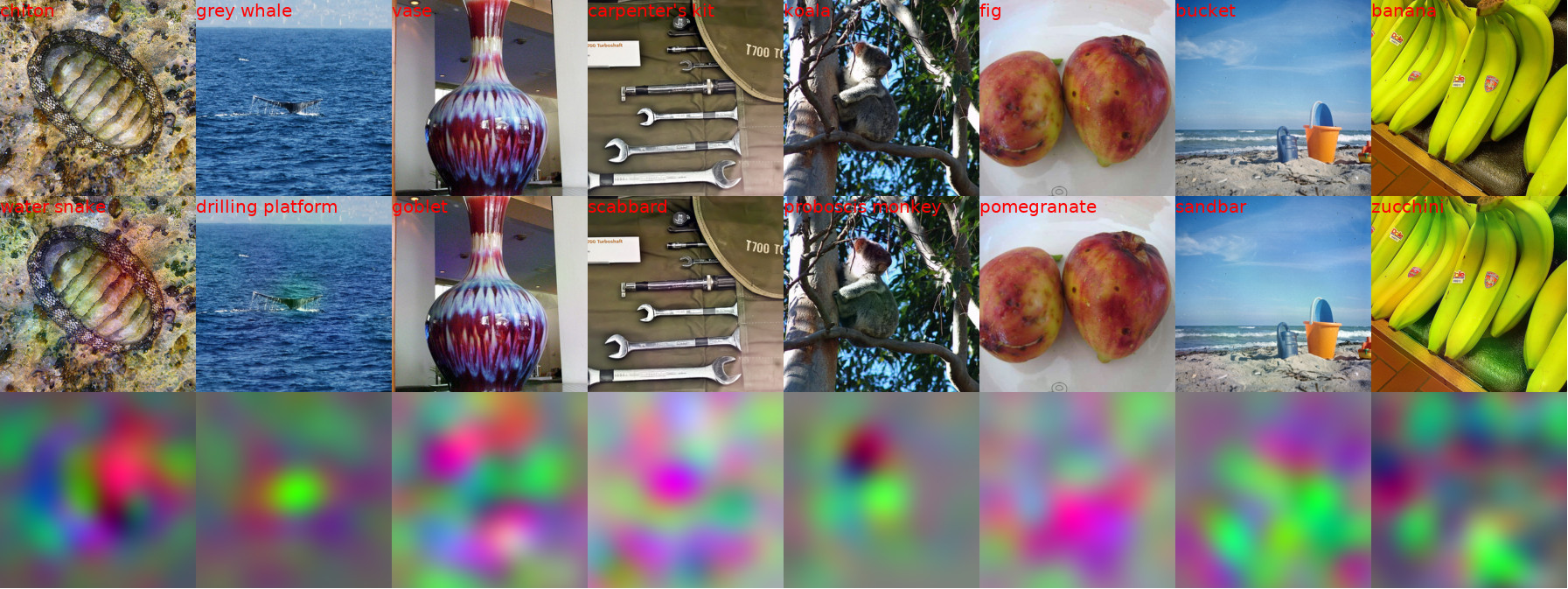}
    \caption{Network architecture: \textbf{ResNet-152}, $\sigma_{\g}=\textbf{20}.$
    }
        \label{fig14}
\end{figure*}

\begin{figure*}
    \includegraphics[width=480pt]{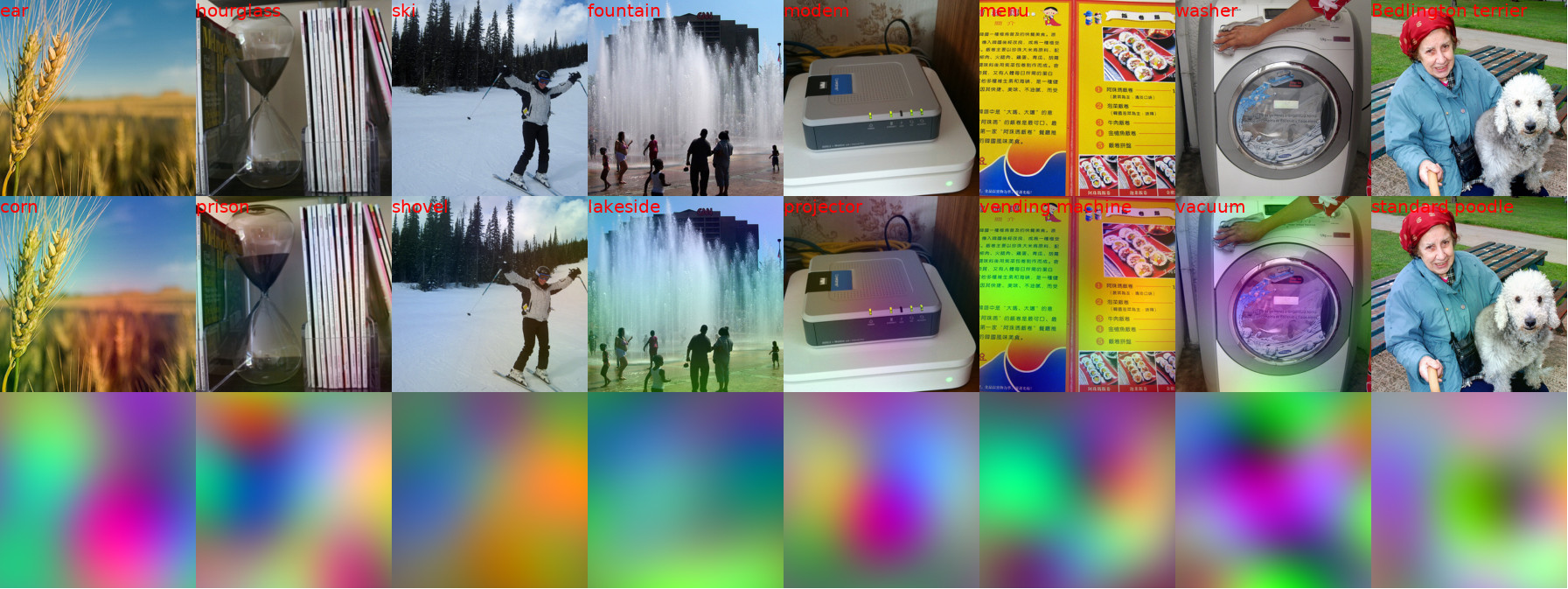}
    \caption{Network architecture: \textbf{ResNet-152}, $\sigma_{\g}=\textbf{50}.$
    }
        \label{fig15}
\end{figure*}

Figure \ref{fig:uaps} demonstrates further smooth UAPs computed for different network architectures. Figures \ref{fig1}-\ref{fig15} show several examples of generated smooth perturbations for images from the validation set of ILSVRC2012 and different network architectures. Each figure demonstrates the results for a fixed value of $\sigma_{\g}$. Each row, from top to bottom, shows benign images, adversarial images, and the corresponding smooth APs, respectively. The predicted label for each image is overlaid on the image in red-colored text. The target network architecture and the smoothing factor $\sigma_{\g}$ for each figure is denoted in the caption. 

\begin{algorithm}
\caption{Computation of smooth UAPs.}
\begin{algorithmic}[1]
\State \textbf{input:} Data points $X=\{\x_1,\dots,\x_m\}$, classifier $f$ and the associated class predictor $c$, desired fooling rate $\tau$, low-pass filter $\g$, desired $\ell_{\infty}$-norm $\xi$.
\State \textbf{output:} Smooth UAP $\v$.
\State Initialize $v \gets 0$.
\While{$\dfrac{1}{m}\sum\limits_{i=1}^{m}\mathds{1}_{>0}{(|c(\x_i) - c(\x_i+v)|)}<\tau$}
\For{each $i \in \{1,\dots, m\}$}
\If{$c(\x_i+\v) = c(\x_i)$}
\State $\r_i = \text{SmoothFool}(f, \x_i, \g)$
\State $\v \gets \v+\r_i$
\State $\v \gets (\v/\xi+1)*0.5$
\State $\v \gets \text{SmoothClip}(\textbf{\underline{0}}, \v, \g, \epsilon)$
\State $\v \gets (2*\v-1)*\xi$
% \State $$
% \State $\r_i \gets \text{clip}(\v+\r_i, \xi) - \v$
\EndIf
\EndFor
\EndWhile
\State \textbf{return} $\v$.
\end{algorithmic}
\label{alg1}
\end{algorithm}


\end{document}